\documentclass[journal]{IEEEtran}
\usepackage[utf8]{inputenc}
\usepackage[T1]{fontenc}
\usepackage{booktabs}
\usepackage{bm}
\usepackage{caption}
\usepackage{subcaption}
\usepackage{multirow}
\usepackage{makecell}
\usepackage{epsfig}
\usepackage{graphicx}
\usepackage{amsmath}
\usepackage{amssymb}
\usepackage{multirow}
\usepackage{color}

\begin{document}

\title{Scene Recognition with Objectness, Attribute and Category Learning}

\newcommand{\comment}[1]{}
\newcommand{\etal}{et al.} 
\newcommand\numberthis{\addtocounter{equation}{1}\tag{\theequation}}

\author{Ji~Zhang, ~\IEEEmembership{Fellow,~IEEE,}
        Jean-Paul~Ainam, ~\IEEEmembership{Member,~IEEE,}
        Li-hui~Zhao, 
        Wenai~Song,
        and~Xin~Wang
\thanks{Ji Zhang, J.-P Ainam, W. Song are with the School
School of Software, North University of China, Taiyuan, China}
\thanks{X. Wang is with Southwest Petroleum University, China.}
\thanks{W. Song is with Shanxi Research Center of Software Engineering Technology, Taiyuan, China}
}

\maketitle

\begin{abstract}
Scene classification has established itself as a challenging research problem. Compared to images of individual objects, scene images could be much more semantically complex and abstract. 
Their difference mainly lies in the level of granularity of recognition. Yet, image recognition serves as a key pillar for the good performance of scene recognition as 
the knowledge attained from object images can be used for accurate recognition of scenes. 
The existing scene recognition methods only take the category label of the scene into consideration. However, we find that the contextual information that contains detailed local descriptions are also beneficial in allowing the scene recognition model to be more discriminative. 
In this paper, we aim to improve scene recognition using  attribute and category label information encoded in objects. Based on the complementarity of attribute  and category labels, we propose a Multi-task Attribute-Scene Recognition (MASR) network which learns a category embedding and at the same time predicts scene attributes. Attribute acquisition and object annotation are tedious and time consuming tasks. We tackle the problem by proposing a partially supervised annotation strategy in which human intervention is significantly reduced. The strategy provides a much more cost-effective solution to real world scenarios, and requires considerably less annotation efforts. Moreover, we re-weight the attribute predictions considering 
the level of importance indicated by the object detected scores.
Using the proposed method, we efficiently annotate attribute labels for four large-scale datasets, and systematically investigate how scene and attribute recognition benefit from each other. The experimental results demonstrate that MASR learns a more discriminative representation and achieves competitive recognition performance compared to the state-of-the-art methods.
\end{abstract}


\IEEEpeerreviewmaketitle

\section{Introduction}\label{sec:introduction}

\IEEEPARstart{S}{cene} recognition, a.k.a, scene categorization, aims to determine the overall scene category (e.g., beach, kitchen, airport) by laying an emphasis on understanding its global properties \cite{Cordts2016CityscapeDataset, Zhou2014PlacesDataset}. It is a high-level computer vision task that allows definition of context for object recognition. Whereas tremendous progress in object detection and semantic segmentation tasks has been achieved \cite{Alejandro2020SemanticAware, Wu2019detectron2}, the performance at scene recognition has not attained the same level of success \cite{Cordts2016CityscapeDataset, Zhou2017ADE20KDataset}. Despite the increasing attention given by researchers to solve the scene recognition problem, it still remains a challenging task.

Early attempts at recognizing high-level scene properties used hand-engineered features \cite{Xiao2010SunDatabase}. For instance, Xiao \etal \cite{Xiao2010SunDatabase} investigated the benefits of several well-known low-level descriptors such as HOG \cite{Dalal2005HOG}, SIFT \cite{Lowe1999SIFT}, and SSIM \cite{Shechtman2007SSIM} while Meng \etal \cite{Meng2012SceneRecognition}  proposed a global image representation using Local Difference Binary Pattern (LDBP).
These methods can learn sophisticated features to describe the visual appearance of scenes. However, they rely heavily on specific types of visual cue, such as color, texture, or shape, which are not practical and powerful enough to discriminate scenes with similar visual appearances.

Current techniques for assigning semantic label to scenes are mostly based on Convolutional Neural Networks (CNNs) \cite{Cimpoi2015DeepFilter, Donahue2014DeCAF, Xie2017HybridCNN}. Specifically, Donahue \etal \cite{Donahue2014DeCAF} investigated the semantic clustering of deep convolutional features, and Xie \etal \cite{Xie2017HybridCNN} proposed to combine CNN features with dictionary-based models. These methods only use the image features 
 to recognize the scene and fail to produce robust scene representations. 
Other recent works consider semantic information about object attributes and states. 
For example, Chen \etal \cite{Chen2020PrototypeAgnostic} adopted a contextual-based model for discovering regions in scene images using spatial structural layouts and Alejandro \etal \cite{Alejandro2020SemanticAware} used semantic segmentation as an additional modality of information for scene recognition. 
\begin{figure}
  \centering
  \includegraphics[width=\linewidth]{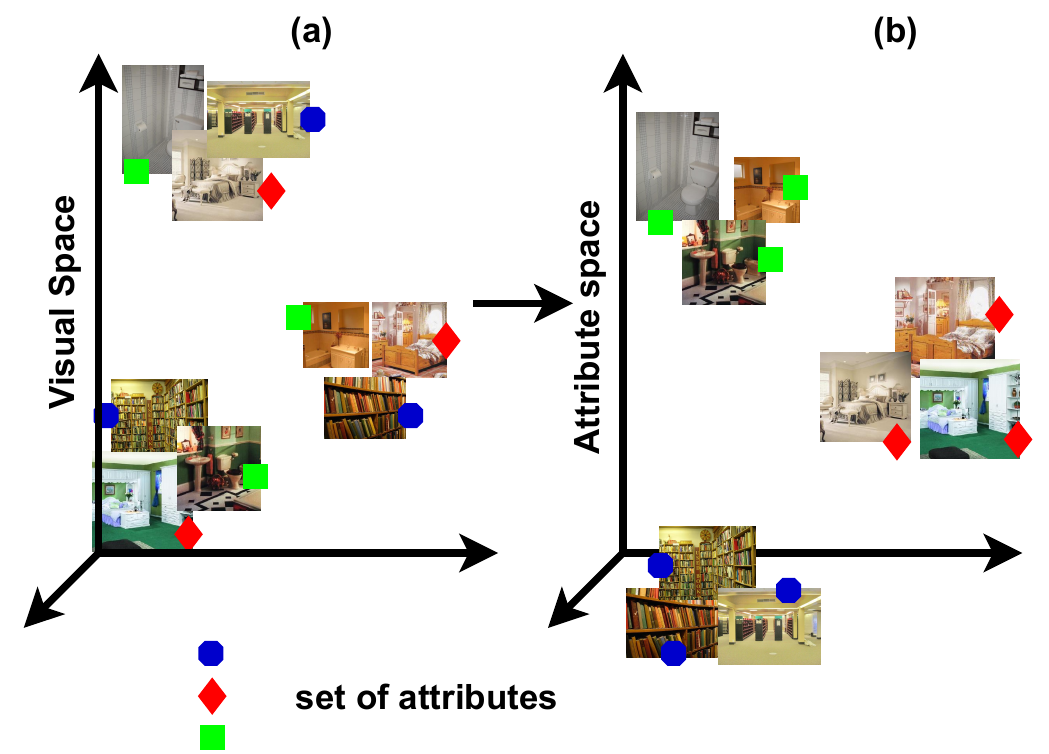}
  \caption{Motivation: using visual feature alone, it's difficulty to discriminate visually similar images. On the other hand, attributes are semantically descriptive across classes. 
  }\label{fig:motivation}
\end{figure}


In general, context information such as semantic segmentation, structural layouts and object attributes have become the key to improve scene recognition accuracy.
In particular, semantic attributes are used to enable a richer description of scenes 
while semantic segmentation
 enables spatial relationships between objects in a scene. 
As shown in Figure \ref{fig:motivation}, it is difficult to distinguish the three categories using only visual features, while on the attribute space, it's easier to semantically differentiate the images across categories.  Attribute information are quite important to discriminate similar images and boost the performance of scene recognition. 
However, extracting object attributes or building an effective semantic representation have proven to be quite challenging, especially when the object attribute annotation must be made through human effort. Semantic segmentation is also challenging given that the task of labeling a scene with accurate per-pixel labels is time consuming.

\begin{figure}
  \centering
  \includegraphics[width=\columnwidth]{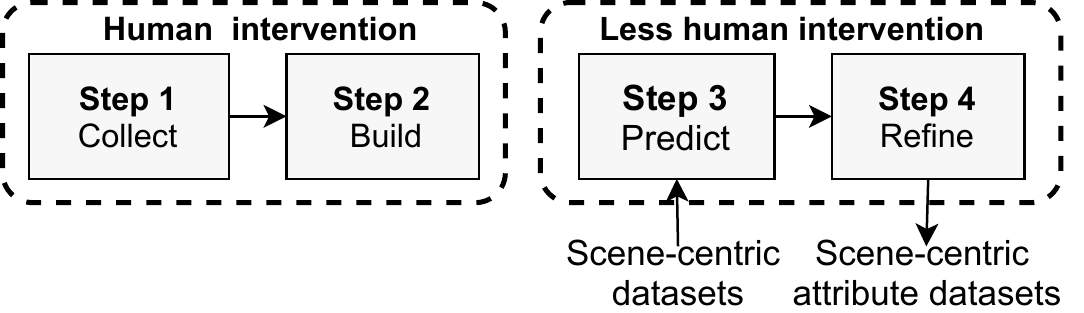}
  \caption{The four steps of our annotation strategy. We create a new set of object attributes for a scene dataset using predicted objects from object detection models. 
  }\label{fig:annotationstrategy}
\end{figure}

To overcome the above limitations, we first propose a partially supervised annotation strategy in which human supervision and intervention are reduced and are only required on few tasks. The strategy is flexible and provides a much more cost-effective solution to real world scenarios, and requires much less annotation efforts.
An overview of the annotation approach is depicted in Figure \ref{fig:annotationstrategy}. We begin by identifying pre-trained models for the task of object detection. 
Then, we use these models to predict object instances for the scene images. Since image scenes are composed of objects, we can use object predictions as attributes to describe scenes. However, this simple, yet effective approach has a major drawback in that attribute predictions are not always highly reliable due to possible biases on their training data. We solve this problem by only keeping those attributes with high confidence scores while discarding those attributes with low confidence scores. In addition, we include these attributes into our model through a regularized loss specifically adapted for this purpose.
  Moreover, Zamir \etal \cite{Zamir2018Taskonomy} showed that scene recognition, object detection and semantic segmentation are interrelated tasks that share a common Branch and that, under transfer learning setting, the performance of object detection methods is generally higher than that of scene recognition. As a result, objects predictions mined from scene images using object detection models can also benefit the scene recognition task.

Following the strategy above, we exploit both local (attribute features) and global information (image features) to learn a better scene representation. We use the detected score to control the importance of each attribute while previous works based on attributes do not consider their importance. Many attributes usually co-occur for a scene with a different level of importance, and these scores may be helpful. 
For example, the object ``table", ``computer'' and ``chair'' better describe the scene category `office'' than the object 
``suitcase". Given these objects in a scene, the contributions of ``table", ``computer'' and ``chair'' must be considered. Motivated by this, we therefore introduce a Re-weighting Attribute Layer that utilizes the confidence scores of object in the scene to optimize scene recognition.

 Compared with previous methods, our work differs in two aspects. First, we propose to investigate object information from object-centric datasets as attribute labels to the task of scene recognition. These attribute labels provide detailed local descriptions.
Secondly,  we propose a Multi-task Attribute-Scene Recognition (MASR) network that exploits attribute information in terms of objects scores to improve the scene recognition task. Various object and context information learned from object detection models are utilized as attributes. Moreover, we categorize the list of attributes into the probable and significant ones based on their prediction effectiveness 
and introduce a layer which adaptively incorporates the attribute into the network.

Our contributions are summarized as follows:
\begin{enumerate}
  \item We propose an attribute mining strategy and release a set of scene attributes for MIT67 \cite{Quattoni2009MIT67Dataset}, SUN397 \cite{Xiao2010SunDatabase}, ADE20K \cite{Zhou2017ADE20KDataset} and Places-365 \cite{Zhou2018PlacesDataset} datasets. Code and attribute annotations for the datasets will be made publicly available after the blind-review process is completed. 
  \item We propose a novel multi-task attribute-scene recognition framework. The framework learns discriminative CNN embeddings for both scene recognition and attribute recognition.
  \item We introduce an attribute re-weighting layer which leverage the attribute predictions to correct the classification based on the detected scores. 
  \item We achieve very competitive results for our proposed the method compared with the state-of-the-art scene recognition methods on four large-scale datasets. 
\end{enumerate}

The remainder of this paper is organized as follows. In Section \ref{sec:relatedworks}, we present the related research work. Then, in Section \ref{sec:annotationstrategy}, we describe our attribute annotation strategy. In Section \ref{sec:ourapproach}, we provide details of the proposed scene recognition model. In Section \ref{sec:experiments}, we present our experimental results on scene recognition task. Finally, we conclude this paper and discuss the future works in Section \ref{sec:conclusion}.

\section{Related Works} \label{sec:relatedworks}
In this section, we describe the existing works that are relevant to our approach. We start by presenting works for solving the scene recognition problem in general, then we elaborate on multi-modal approaches and, finally, we present the recognition models that make use of contextual information.

\subsection{Scene recognition}
Research for scene recognition can be roughly divided into two groups: those based on handcrafted feature representations and those based on deep machine learning. The first group of works focus on 
designing appropriate local descriptors for scenes and the second group uses CNN to learn scene embeddings.

Hand engineered descriptors such as SIFT \cite{Lowe1999SIFT}, HOG \cite{Dalal2005HOG} have been the fundamental component for computer vision tasks, including scene recognition. Bag-of-features (e.g. VLAD 
, Fisher kernel) 
have also shown great success on scene recognition. Laurent \etal \cite{Laurent2005GIST} proposed to design a holistic low-level features using Generalized Search Trees (GIST) descriptor. Other works have combined local features from local patches with holistic features. For example, Lazebnik \etal \cite{Lazebnik2006BeyondBags} proposed a technique that partitions the image into fine sub-regions and computes histograms of local features for each sub-region. The resulting spatial pyramid is simply an efficient extension of an orderless bag-of-features image representation. Similarly, Krapac \etal \cite{Krapac2011Modeling} introduced an extension of bag-of-words image representations and Fisher kernels to encode both the spatial layout and the appearance of local features. In addition, Wu \etal \cite{Wu2011CENTRIST} proposed CENsus TRansform hISTogram (CENTRIST), a visual descriptor for scene categorization that encodes the local structural properties within an image and suppresses detailed textual information. However, CENTRIST is not invariant to rotations and only utilizes the gray-scale information of images, which limits its application.
Furthermore, Margolin \etal \cite{Margolin2014OTC} proposed the Oriented Texture Curves (OTC) descriptor to capture the texture of a path along multiple orientation using the shapes of multiple curves, the gradients and curvatures, and the local contrast differences.
In general, these handcrafted methods have failed as they depend on certain types of visual cue (e.g. color, texture, or shape) that are insufficient to distinguish between similar scenes. 
In addition, due to the handcraft nature of the features, these methods suffer a huge data bias without the flexibility on new datasets.

 The second group of methods based on CNN usually result in better performance. Many works 
 combine orderless bag-of-features  or dictionary with CNN features to incorporate discriminative local and structural information. In particular, Cimpoi \etal \cite{Cimpoi2015DeepFilter} used a texture descriptor obtained by Fisher Vector pooling of a CNN filter bank and Xie \etal \cite{Xie2017HybridCNN} combined features extracted from CNN with two dictionary-based representations. 
However, because of to the huge inter-class similarity in scene images, these approaches do not produce scene representations that are robust. 
To overcome this problem, recent techniques are based on multi-modality approach that incorporates context, attribute and object information into CNN to constrain scene recognition.

\subsection{Hybrid and Multi-Modal architectures}

In recent years, researchers have investigated how to effectively integrate local semantics of both objects and concepts in scene classification. In particular, Shuqiang \etal \cite{Shuqiang2019Codebook} proposed to combine a local feature codebook generated from both ImageNet  and Places-365 datasets 
 with the original features from scene images. Generally, hybrid-based methods first extract the local representations from image patches, and then aggregate their local representation by encoding methods. For example, Gong \etal \cite{Gong2014Orderless} proposed a multi-scale orderless approach that performs orderless VLAD pooling on CNN activations at each level separately, and concatenates the resulting local representations. Yang \etal \cite{Yang2015DAGCNNs} explored multi-scale CNNs by extracting features from multiple layers. 
 Similary, Cheng \etal \cite{Cheng2018Objectness} proposed a semantic descriptor where correlation of object configurations across scenes are exploited. ImageNet-CNN output score at the softmax layer is used to compute multinomial distribution using Bayes rule.
 However, Zhou \etal \cite{Zhou2014PlacesDataset} showed that incorporating ImageNet features do not help much and demonstrated that object-centric and scene-centric neural networks differ in their internal representations. As a result, Zhou \etal proposed to learn from massive amounts of data by introducing a new benchmark with millions of labeled images. Similarly, Herranz \etal \cite{Herranz2016MultiScales} presented an alternative method of combining features from an object-centric dataset that uses different scale ranges. Herranz \etal mainly addressed the scale induced dataset bias in multi-scale CNN architecture and showed an effectively way of combining scene-centric and object-centric knowledge (i.e., Places and ImageNet) in CNNs. However, the recognition accuracy highly depends on the scale, and carefully chosen multi-scale combinations are required to push the state-of-the-art recognition.

Recently, Yang \etal \cite{Yang2018Dictionary} proposed a dictionary learning layer composed of a finite number of recurrent units to simultaneously enhance the sparse representation and discriminative abilities of features. Pei \etal \cite{Pei2021PlacePerception} used a mixed CNN-LSTM network that combines both visual and linguistic features like image captions. 
Furthermore, Alejandro \etal \cite{Alejandro2020SemanticAware} exploited the spatial relationship between objects using both RGB images and semantic segmentation. 
However, as shown by the study of \textit{taskonomies} \cite{Zamir2018Taskonomy}, 
the performance of semantic segmentation methods is generally lower than that of object detection approaches. Thus, in this paper, we propose to enhance scene recognition using context information mined from object detection models.

\subsection{Attributes for Image Recognition}
The use of complementary information such as attributes has been proposed in several computer vision tasks including pedestrian recognition \cite{Lin2019AttributeReID, Tay2019AANetReID}, action recognition \cite{Yao2011ActionRecognition, Roy2019UniversalAttribute}, image recognition \cite{Ferrari2007VisualAttributes, Lampert2009AttributeTransfer}, and scene recognition \cite{Zeng2020SceneAttribute, Patterson2012SUN}.

In pedestrian recognition, Lin \etal \cite{Lin2019AttributeReID} proposed a multi-task network which learns an identity embedding and at the same time predicts pedestrian attributes. Tay \etal \cite{Tay2019AANetReID} also proposed an Attribute Attention Network (AANet) that integrates person attributes and attribute attention maps into a unified framework.
The effectiveness of attributes has also been studied in action recognition. In particular, 
Rueda \etal \cite{Rueda2018HumanActivity} introduced a search for attributes that represents signal segments for recognizing human activities 
while Yao \etal \cite{Yao2011ActionRecognition} combined attributes and parts. Here, the attributes are represented as verbs describing human actions, and parts are composed of objects related to the actions. Recently, Roy \etal \cite{Roy2019UniversalAttribute} used local spatio-temporal features to capture the action attributes in a Guassian mixture model.

In image recognition, basic attributes such as texture, shape and color  have been extensively investigated in early works \cite{Ferrari2007VisualAttributes, Lampert2009AttributeTransfer}. In particular, Ferrari \etal \cite{Ferrari2007VisualAttributes} showed for the first time that attributes can be learned for object recognition through weakly supervised learning and trained a set of classifiers to predict the existence of human-labeled attributes in the data. Both Lampert \etal \cite{Lampert2009AttributeTransfer} and Xian \etal \cite{Xian2018AWA2Dataset} proposed transfer learning and zero-shot learning approaches respectively on the Animals with Attributes datasets (AwA and AwA2) \cite{Xian2018AWA2Dataset, Lampert2009AttributeTransfer}. The two datasets contain $50$ classes and $85$ attributes with no image overlap.


In scene recognition, there exists only few works about attributes. For instance, Oliva \etal \cite{Olivia2001Holistic} proposed a small-scale scene dataset with $8$ attributes describing the spatial structure of a scene (e.g., naturalness, openness, roughness etc.) and Wang \etal \cite{Wang2013AttributeLocalization} created an outdoor scene dataset with $47$ attributes.  
In addition, Patterson \etal \cite{Patterson2012SUN} introduced a subset of the SUN dataset \cite{Xiao2010SunDatabase} containing $14,340$ images annotated with $102$ attributes and used the attributes as mid-level semantic information for scene recognition. Recently, Zeng \etal \cite{Zeng2020SceneAttribute} proposed to aggregate more complementary visual features of the scene using 
features from the Attribute-ImageNet \cite{Ouyang2015AttributeImageNet} and the Places \cite{Zhou2014PlacesDataset} datasets.

In contrast to existing methods using attributes in general \cite{Ferrari2007VisualAttributes, Lampert2009AttributeTransfer, Roy2019UniversalAttribute, Xian2018AWA2Dataset, Zeng2020SceneAttribute}, and scene attributes in particular \cite{Olivia2001Holistic, Patterson2012SUN}, we do not manually annotate the attributes and we do not directly learn a weight for each attribute to control the attribute's impact, but we leverage object detected score information contained in the attributes. By so doing, the model could use more of its parameters for learning to compensate in failure cases.
 Moreover, we go beyond the simple use of objects in scene recognition and propose to utilize existing object-centric datasets to bridge the gap between object recognition and scene recognition.


\section{Attribute annotation strategy} \label{sec:annotationstrategy}
 Most datasets for scene recognition \cite{Cordts2016CityscapeDataset,Zhou2017ADE20KDataset, Xiao2010SunDatabase, Quattoni2009MIT67Dataset, Zhou2018PlacesDataset} provide labels that rarely fill the gap between scene classification and semantic scene description. Recently, an effort is made to provide not only category labels to scene, but also attribute labels for describing the objects within the scene \cite{Patterson2012SUN, Olivia2001Holistic, Wang2013AttributeLocalization}. However, the intensive human labor required to annotate such datasets limits its application, especially when the attribute distribution is large. In this paper, we propose to mine object and context information using object detection models. In particular,
we start by collecting object attributes and context information from two popular object-centric datasets, namely COCO Object and COCO Panoptic, and then use the attributes for scene classification after discarding the less discriminative ones. COCO Object is an object-centric dataset that provides label to $80$ different objects (i.e., \textit{things}) while COCO Panoptic is an instance segmentation task addressing both \textit{stuff} and \textit{thing} classes. Here, \textit{things} are countable objects (e.g., people, animals etc.) whereas \textit{stuff} are regions of similar texture or material (e.g., sky, grass, road etc.). \textit{Stuff} and \textit{thing} classes are usually addressed independently and included separately to boost scene recognition \cite{Alejandro2020SemanticAware, Cheng2018Objectness}. 
In this paper, we use both \textit{stuff} and \textit{thing} with scene categories in a unified framework. Table \ref{tab:stuffandthing} presents some class instances for \textit{stuff} and \textit{thing}.

\begin{table}
  \centering
  \caption{Examples of stuff and thing classes from COCO Object and COCO Panoptic datasets.}
  \label{tab:stuffandthing}
  \begin{tabular}{|l|l|}
    \hline
    Groups & Attributes \\
    \hline
    \textit{Things} & bottle, cup, apple, sheep, dog, suitcase, tv, toilet...\\
    \hline
    \textit{Stuff} & sea, river, road, sand, snow, wall, window, wall...\\
    \hline
  \end{tabular}
\end{table}

Our annotation strategy can be formalized as follow. Let $S$ and and $T$, be the set of \textit{stuff} and \textit{thing}, respectively. $F_s$ and $F_t$, two pre-trained CNN models for the respective tasks. We denote by $\{x_1, x_2, \ldots, x_n \} \in X$ a scene-centric dataset with only category labels. Our goal is to use $F_s$ and $F_t$ on $X$ to predict distributions over $S$ and $T$ 
such that,
\begin{equation}\label{eq:ptps}
p_s  = F_s(X) \quad \text{ and } \quad p_t = F_t(X)
\end{equation}
where $p_s \in \mathbb{R}^{\vert S \vert}$ and $p_t \in \mathbb{R}^{\vert T \vert}$ are the probability predictions over $S$ and $T$, respectively.
Given $X$, the final \textit{stuff} + \textit{thing} predictions $\mathcal{P} \in \mathbb{R}^{\vert S \vert + \vert T \vert}$, on the given scene dataset is defined as
\begin{equation}\label{eq:psunionpt}
  \mathcal{P} = p_s \cup  p_t
\end{equation}
Note that, $\mathcal{P}$ does not add to $1$ and does not represent a probability distribution. A common method of consolidating the two probability distributions $p_s$ and $p_t$ is to simply average them for every set of values $V$, s.t. $\mathcal{P}(V) = \frac{p_s(V) + p_t(V)}{2}, \quad \forall V \in S \cap T$. However, here, $S$ and  $T$ do not always overlap and usually describe different data sources. Consequently, we simply merge $p_s$ and $p_t$ and use the object detection score as a confidence score.   

\begin{table}
  \centering
    \caption{Some scene categories with the total number of attributes (\#Att), the number of attributes discarded using Equations \ref{eq:discard}, \ref{eq:discard2} and the number of attributes used (\#Used) for the MIT67 dataset.}\label{tab:attributepercategory}
  \begin{tabular}{|l|c|c|c|c|}
    \hline
    Categories & \#Att & \# Eq. \ref{eq:discard} & \# Eq. \ref{eq:discard2} & \#Used \\
    \hline
    airport\_inside & 51 & 12 & 39 & 6 \\
    cloister & 31 & 8 & 23 & 5 \\
    computerroom & 48 & 12 & 36 & 8 \\
    dining\_room & 43 & 7 & 36 & 9 \\
    fastfood\_restaurant & 63 & 16 & 47 & 7 \\
    waitingroom & 37 & 8 & 29 & 6 \\
    office  & 49 &  8 &  41 & 9 \\
    bathroom & 45 & 14 & 31 & 10 \\
    \hline
  \end{tabular}
\end{table}

\begin{figure*}
  \centering
  \includegraphics[width=0.8\linewidth]{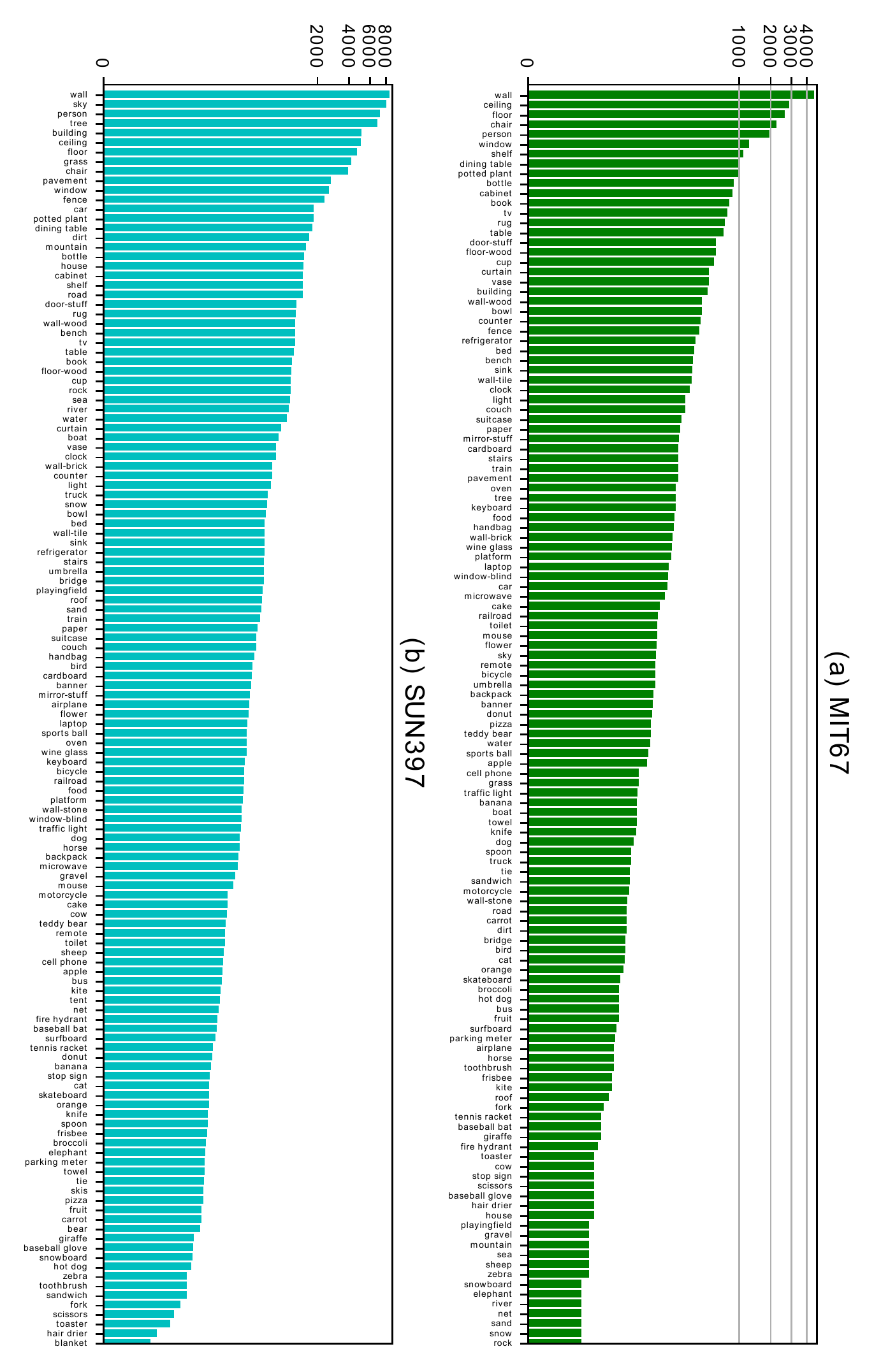}
  \caption{Attribute distribution sorted by frequency on MIT67 and SUN397 datasets. (Zoom in for best view).}\label{fig:frequenceinstance}
\end{figure*}

\subsection{Filtering less discriminative attributes}
   Generally, scene categories such as `airport-inside' and `offices' contain slightly more objects than other places (See Table \ref{tab:attributepercategory}). This prior knowledge - also inferred from Figure \ref{fig:frequenceinstance}) - can lead to a common problem in most scene recognition methods that use objects as a clue. When there is too much information such as attributes and relationships about the objects, it may be challenging to identify the category of the place correctly. Since our strategy selects several information from object-related tasks, it can not guarantee that all of the selected objects are helpful for scene recognition. To overcome this problem, we propose to further select elements of $S$ and $T$ based on their frequencies and detection scores. 

       \textbf{Based on the detection scores}: The object instances with detection score less than a threshold are discarded. Only objects with scores higher than a threshold are selected as scene attributes. We think that narrowing the scene description to the list of probable objects is much more realistic that using all the detections. We 
       redefine $\mathcal{P}$ as :
\begin{equation}\label{eq:discard}
  \mathcal{P}^* = \bigg\{ p_i \bigg\}_{i=1}^{\vert \mathcal{P} \vert} \quad \bigg\vert \quad  p_i \in \mathcal{P} \text{ and } p_i > \xi
\end{equation}
where $\xi$ is the threshold. 
When the detection score is 0, the object is considered not present in the scene.

\textbf{Based on the object frequency}:
We further consider attribute frequencies given a scene category and remove less common objects. For each category $c$, we define a relative attribute frequency as the number of non-zero score covering the category images. If $\{ a^1, a^2, \ldots, a^m \} \in \mathcal{A}_c$ is the set of detected attributes for $c$, an optimal $\mathcal{A}^*_c$ is defined as
\begin{equation}\label{eq:discard2}
  \mathcal{A}^*_c = \{ a^j, \quad \text{s.t} \quad f_c(a^j) \geq \beta \}_{j=1}^m
\end{equation}
where $f_c(a^j)$ is the relative frequency of attribute with the value $a^j$ given the category $c$, $\beta$ is the minimum frequency and $\mathcal{A}_c^* \in \mathbb{R}^{1 \times m}$ is the final list of attributes for $c$. As shown in Table \ref{tab:attributepercategory}, using Equations \ref{eq:discard} and \ref{eq:discard2}
 we can be sure to choose only representative objects to describe a category.



\subsection{Comparison and dataset statistics}
Many existing attribute datasets used `yes/no' (or 0/1) to indicate whether the attribute is present or absent in the image \cite{Lin2019AttributeReID, Patterson2014BeyondCategories}.
Even though this worked well and achieved some improvement in scene recognition task, it does not reflect how the humans often describe scene.  
Instead, in this paper, we use the detected score associated with the objects occurring in the image, mimicking a more natural object occurrence in daily scene. Moreover, the detected set of attributes can be easily and continuously enlarged as object detector models get more robust and efforts are made to annotate large object-centric datasets. A scene image is finally represented as a bag of attributes with their respective confidence scores. In Table \ref{tab:samples}, we show some detected objects with their scores.
 
\begin{figure}
  \centering
  \includegraphics[width=\linewidth]{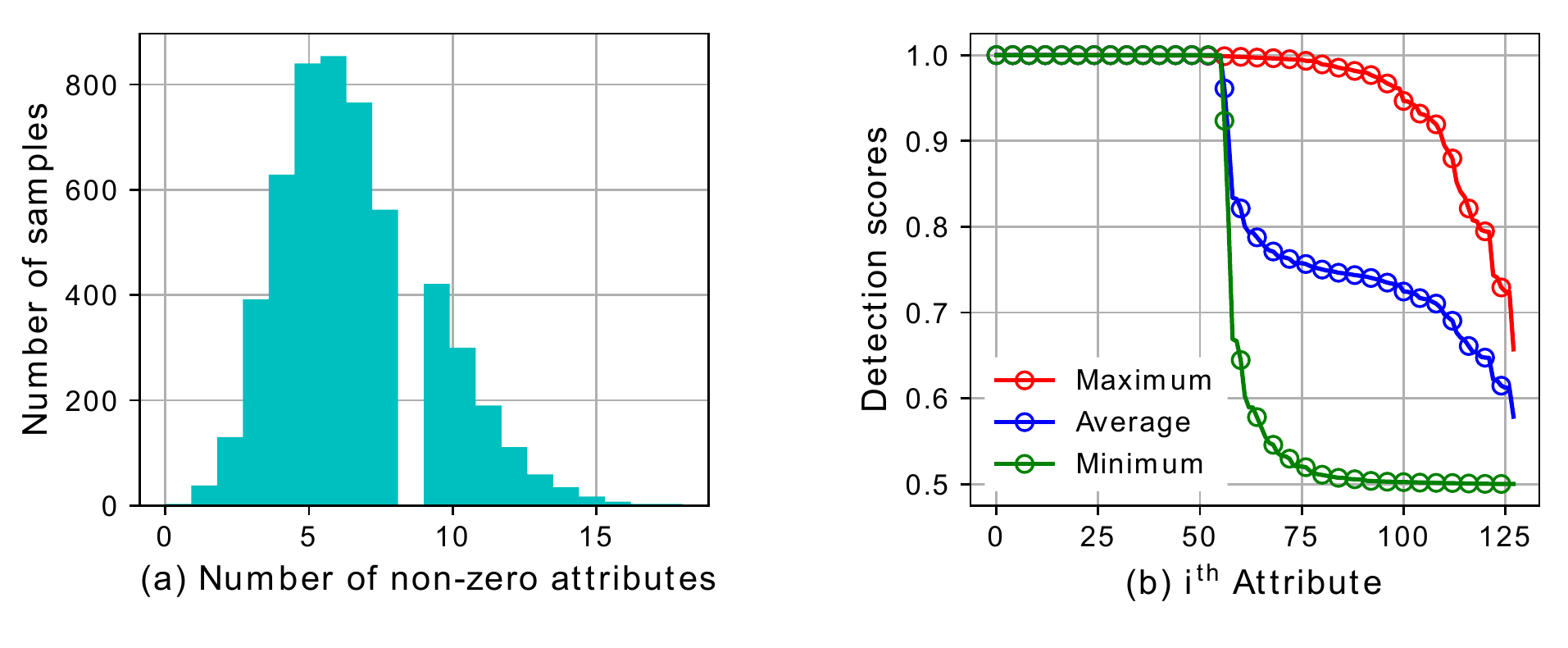}
  \caption{Statistics from MIT67 dataset. (a) The histogram of samples per predictions. (b) The maximum, average and minimum detection scores for $125$ objects.}\label{fig:detectionscore}
\end{figure}

 For both MIT67 and SUN397 datasets, we illustrate the attribute distributions in Figure \ref{fig:frequenceinstance}.
Figure \ref{fig:detectionscore}a) shows the number of detected scores per samples. On  average, each image is described by $6.5$ objects present in the scene.
Figure \ref{fig:detectionscore}b) shows the distribution of the detection scores for $125$ different objects. The maximum, average and minimum scores are illustrated. This shows that more than $50\%$ of the objects can describe a scene with a maximum score of $100\%$ and less than $25\%$ produced a minimum score of $50\%$.  On average, $70\%$ to $80\%$ objects from $
\mathcal{S}$ and $\mathcal{T}$ are detected on the MIT67 dataset.


\begin{table*}
    \centering
    \caption{Samples from Detectron2 model \cite{Wu2019detectron2}. We only show objects with more than 80\% precision. Images come from MIT67 andd SUN397 datasets. (Zoom in for best view)}\label{tab:samples}
    \begin{tabular}{ll|ll}
         bathroom &  & fastfood & \\
        \includegraphics[width=0.35\textwidth]{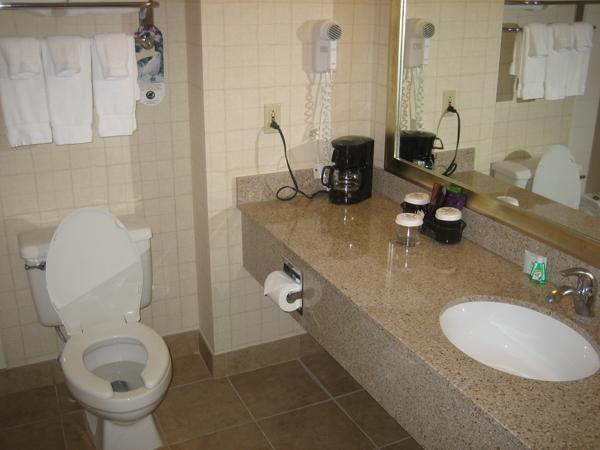}  & 
        \multirow[t]{2}{*}{\makecell{mirror-stuff \\ window \\ 
        toilet \\ floor \\  sink 100\% }
        }& 
        \includegraphics[width=0.35\textwidth]{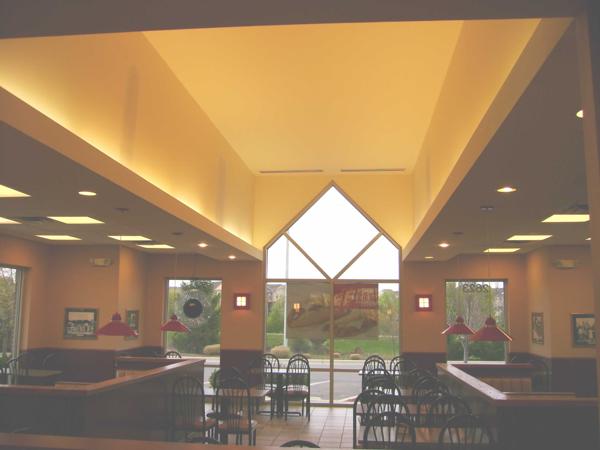} & 
        \multirow[t]{2}{*}{\makecell{
        ceiling \\  wall \\ window \\ chair 98\% \\ dining table
        }}
        \\
        \includegraphics[width=0.35\textwidth]{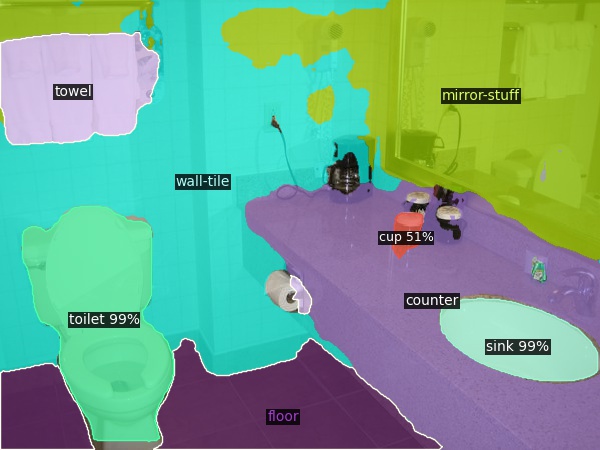} &  & 
        \includegraphics[width=0.35\textwidth]{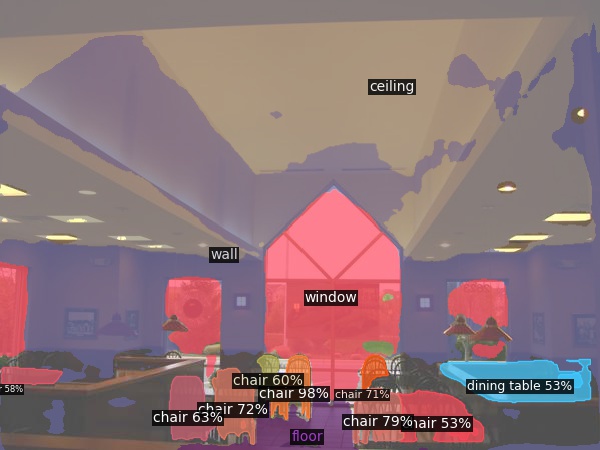} &  \\
        \hline
        airport & & library & \\ 
    \includegraphics[width=0.35\textwidth]{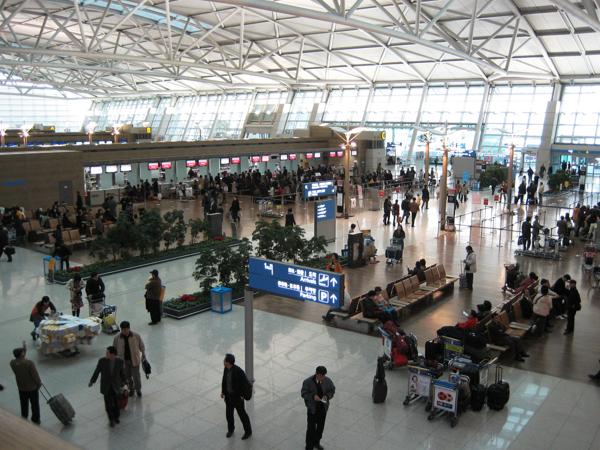} & 
    \multirow[t]{2}{*}{\makecell{
      ceiling \\  wall \\ window \\ person 99\% \\ suitcase 99\%
    }} & 
     \includegraphics[width=0.35\textwidth, height=43mm]{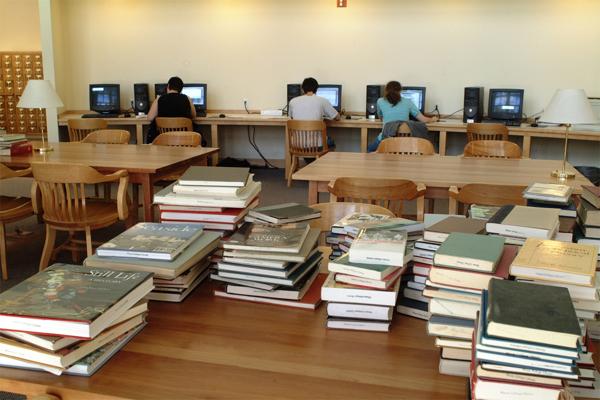} & 
     \multirow[t]{2}{*}{\makecell{
        wall \\ book 97\%  \\ person 97\% \\ chair 97\% \\  tv 94\%
     }} \\ 
     \includegraphics[width=0.35\textwidth]{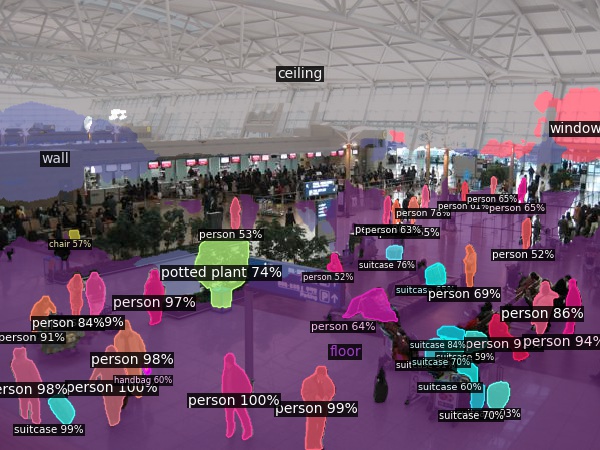} & & 
     \includegraphics[width=0.35\textwidth, height=43mm]{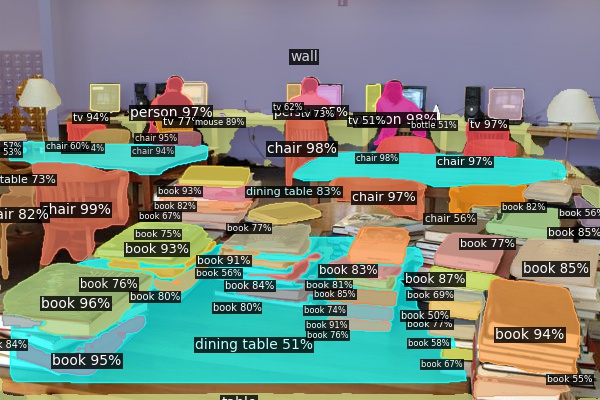} \\
    \end{tabular}
\end{table*}

The proposed strategy can be generalized to $N$ tasks that share a common Branch with scene recognition task. 
Equation \ref{eq:psunionpt} can then be formulated as:

\begin{equation}\label{eq:generalization}
  \mathcal{P} = p_1 \cup p_2 \cup \ldots \cup p_N
\end{equation}
where $p_i = F_i (X)$, $F_i$ a pre-trained CNN model for a task $i$. $\mathcal{P}^*$ and $\mathcal{A}^*_c$ are similarly obtained using Equations \ref{eq:discard} and \ref{eq:discard2}. 
In the next section, we present a method that leverages the scene category and our annotation to improve scene recognition task.

\section{The proposed method} \label{sec:ourapproach}
We start by describing two baselines in Section \ref{sec:baselines}, and then introduce our proposed Multi-task Attribute-Scene Recognition network in Section \ref{sec:MASR}. 
\subsection{Baselines} \label{sec:baselines}
Let $\mathcal{X}_I = \{ (x_1, y_1) \ldots , (x_n, y_n) \}$ be the set of scene images, where $x_i$ denotes the $i$-th image and $y_i$ its corresponding category. For each image $x_i$, we have a set of attribute annotations ${\bf a_i} \in \mathbb{R}^{1 \times m}\quad \vert  \quad {\bf a_i} = \{ a^1_i, a^2_i, \ldots, a^m_i \}$ where $a^j_i$ is the $j$-th attribute label for the image $x_i$, and $m$ is the number of attribute labels. Let $\mathcal{X}_A = \{ (x_1, a_1), \ldots, (x_n, a_n) \} $ be the set of attributes. Note that, $\mathcal{X}_I$ and $\mathcal{X}_A$ share common scene images $\{ x_i \}$. Based on $\mathcal{X}_I$ and $\mathcal{X}_A$, we define two baselines:

\textbf{Baseline 1: Scene classification baseline}.
Similar to \cite{Alejandro2020SemanticAware}, we take different backbones (e.g. ResNet, VGGNet, etc.), from which we modify the classifier layer to match the number of category $K$.  Using only $\mathcal{X}_I$, this baseline considers scene recognition training process as an image classification task and defines an objective function such that:
\begin{equation}\label{eq:classificationloss}
 \min_{\theta_I, w_I} \sum_{i=1}^{n} \ell (f_I(w_I, \phi(\theta_I, x_i)), y_i)
\end{equation}
where $\phi$ is the backbone model parameterized by $\theta_I$ to extract features from the input image $x_i$, $f_I$ is the classifier layer parameterized by $w_I$, which classifies the feature embeddings $\phi(\theta_I, x_i)$ into $K$-dimensional vectors representing the scene categories. $\ell$ is the cross-entropy loss.

\textbf{Baseline 2: Scene attribute baseline}. This baseline uses the attribute data $\mathcal{X}_A$ and predicts the scene attributes. The model is trained using the following objective function:
\begin{equation}\label{eq:baseline2loss}
   \min_{\theta_A, w_A} \sum_{i=1}^{n} \sum_{j=1}^{m} \ell ( f_{A_j}(w_{A_j}, \phi(\theta_A, x_i)), a_i^j)
\end{equation}
where $f_{A_j}$ is the $j$-th attribute classifier, parameterized by $w_{A_j}$, to predict the $m$ attributes using the embedding $\phi(\theta_A, x_i)$. We take the sum of all the losses for $m$ attribute predictions on the input image $x_i$ as the loss. 

For the evaluation of the classification baseline, we use Top@k accuracy metric with $k \in [1, K]$. Top@k measures the percentage of testing data whose topk-scored class coincides with the ground-truth label. To evaluate the attribute baseline, we take the classifier layer output and evaluate it with the ground-truth using the classification metric.

\subsection{Multi-task attribute-scene architecture}\label{sec:MASR}
In this section, we aim to improve scene recognition task using complementary cues from attribute labels. It has been established that scene attributes may contain information which is often highly relevant to the scene recognition task. However, different from existing works, this paper leverages automatically generated attributes with a varying degree of scores. 
\begin{figure}
  \centering
  \includegraphics[width=\linewidth]{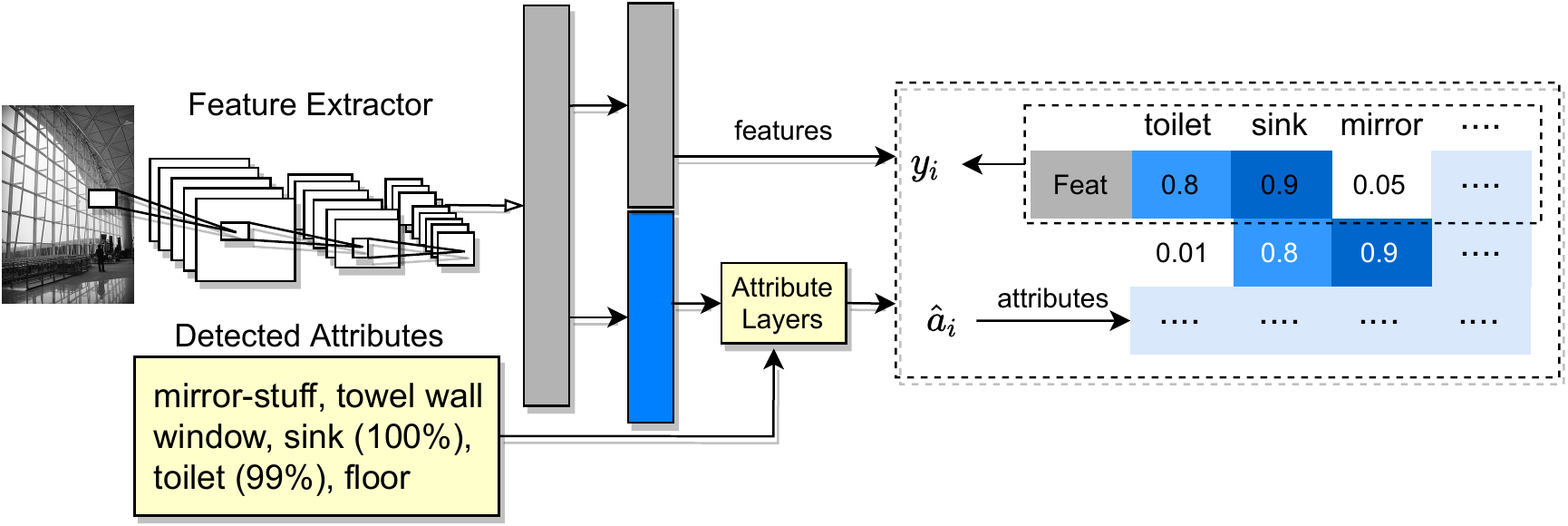}
  \caption{Overview of the MASR architecture. The attribute information are obtained from pre-trained object-centric models. Attributes are used to support the learning of the CNN feature through a regularized loss and a re-weighting layer. The re-weighting layer enforces weights for attributes and reduce the influence of unreliable attributes on the scene result.
   }\label{fig:framework}
\end{figure}
 The \comment{basic} idea is to add a simple, yet effective attribute scores to the category level features such that the attribute information is able to drive the learning procedure. The overview of the proposed approach is illustrated in Figure \ref{fig:framework}. The network contains two parts. The first part predicts the scene category label, and the second part predicts the attribute labels. Given a scene image $x_i$, we first extract its feature representation $\bm{v}_i = \phi(\theta_I, x_i)$, using a CNN-like network. Secondly, we take attribute scores as additional cues and re-weight $\bm{v}_i$. Then the resulting features is passed through a fully connected layer $L^{\vert K \vert}$ for predictions. 
 Simultaneously, we use $\bm{v}_i$ to predict the attribute probabilities $p_{att}$ using a fully connected layer $L^{\vert A \vert}$, where $A$ is the set detected attributes.

\textbf{Attribute task loss}. Since our attributes are not completely mutually exclusive, the prediction of multi-attribute is a multi-label classification problem. The structure of the layer that predicts the attributes is different from the traditional single-label classification layer that includes one cost function. In order to predict all attributes, we employ a multi-class cross entropy loss 
defined as:
\begin{multline}\label{eq:attributeloss}
  \mathcal{L}_{att} = - \frac{1}{n} \sum_{i=1}^{n} \sum_{j=1}^{m} \bigg ( \hat{a}_i^j \log(p_{att}(x_i, a^j_i))  + \\
  (1 - \hat{a}_i^j) \log(1 - p_{att}(x_i, a^j_i)) \bigg)
\end{multline}
where $p_{att}(x_i, j)$ is the predicted class probability on the $j$-th attribute of the training sample $x_i$,
and $\hat{a}_i^j \in \{0, 1\}$ is the attribute ground-truth defined as:
\begin{equation}\label{eq:attributehat}
\hat{a}_i^j=\left\{\begin{array}{ll}
1 & \text { if } a_i^j > \xi \\
0  & \text { otherwise }
\end{array}\right.
\end{equation}

 The loss in Equation \ref{eq:attributeloss} usually suffers from imbalances in the training data. Some objects such as ``person'' are much more frequent than others (see Figure \ref{fig:frequenceinstance}), and we cannot simply compensate by data sampling, because attributes co-occur and balancing the occurrence frequency of one attribute will change that of others. To address this, we introduce a regularizer $\beta^j$ which reflects the relative frequency of $j$-th attribute in the training data (i.e., its ratio of positive to negative attribute labels). Equation \ref{eq:attributeloss} then becomes:
\begin{multline}\label{eq:regularizedattributeloss}
\mathcal{L}_{att} = - \frac{1}{n} \sum_{i=1}^{n} \sum_{j=1}^{m} \bigg ( \hat{a}^j_i 
(1 - \hat{a}^j_i) \cdot \beta^j_k \cdot \log \big(1 - p_{att}(x_i, a^j_i) \big) \bigg ), \\
 \beta^j_k = \frac{\vert\vert a^j_k \vert\vert}{\sum_{l=1}^{K} \vert \vert a^j_l \vert \vert}_{l \neq k}
\end{multline}
where $\vert \vert a^j_k \vert \vert$ is the number of samples holding the $k$-th class label of the $j$-th attribute
(i.e, the magnitude of the $j$-th attribute for the $k$-th scene category).
Note that, the classifiers for different attribute features are not shared.

\begin{figure}
\centering
 \begin{subfigure}[b]{0.48\textwidth}
  \centering
  \includegraphics[width=.7\linewidth]{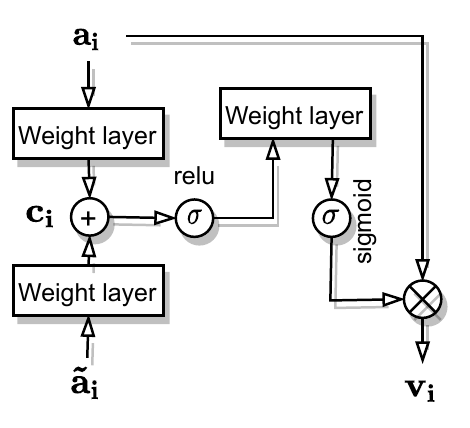}
  \caption{The attribute re-weighting layer (ARL) is applied to the concatenated predictions obtained from each predictions before the sigmoid activation.}
  \label{fig:reweightedlayer}
\end{subfigure}
 \hfill
\begin{subfigure}[b]{0.48\textwidth}
  \centering
  \includegraphics[width=.9\linewidth]{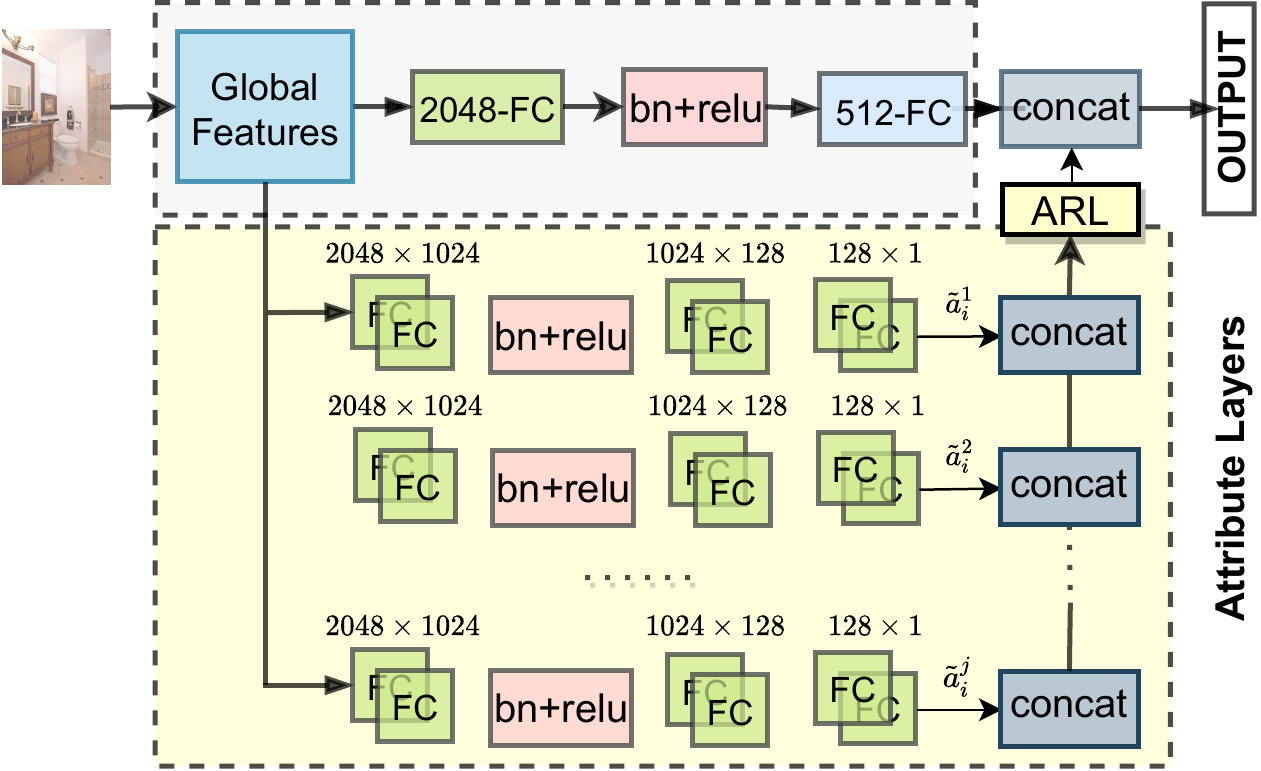}
  \caption{The attribute prediction comprises $j$ multidirectional branches and output $f(a_i^j) \vert a_i^j \in \{a^1_i, a_i^2, \ldots, a^j_i\}$ and $f$ a multi-layer perceptron (MLP). 
  }
  \label{fig:attributeclass}
\end{subfigure}
\caption{The ARL and attribute layers built a cascade of MLP. For each attribute $a^j_i$ of $x_i$, we predict a binary classification.} \label{fig:reweigthingmodule}
\end{figure}

\textbf{Attribute Layers}. Because the attribute representation is learned on separate data, we can expect some of the attributes to be much important that others. We thus introduce a layer that re-weight the attributes using the detected scores. It is composed of some linear transformations that aggregate all attribute information into a single vector ${\bf v_i}$.
We denote by ${\bf\tilde{a}_i} = p_{att}(x_i, {\bf a_i}) \in \mathbb{R}^{1 \times m}$ the attribute scores from the attribute classifier $f_{A}$. We then learn the confidence score ${\bf c_i}$ for its prediction  ${\bf \tilde{a}_i}$ as:


\begin{equation}\label{eq:confidencescore}
  {\bf v_i} = {\bf a_i} * \sigma (W_{c_i} + \underbrace{ \text{ReLU} (W_{a_i} {\bf a_i} + W_{\tilde{a}_i} {\bf \tilde{a}_i} + b_i)}_{\bf c_i})
\end{equation}
where $\sigma$ is the sigmoid activation function, $W_* \in \mathbb{R}^{m\times m}$ and $b_i \in \mathbb{R}^{m \times 1}$ are trainable parameters. The resulting learned parameter $\textbf{c}_i$ is element-wise multiplied with the attribute detection scores ${\bf a_i}$ to produce ${\bf v_i}$. The above operation constitutes the Attribute Re-weighting Layer (ARL). We implement the attribute layers as cascade of ARL layers.  Finally, ${\bf v_i}$ is concatenated with the global image representation for further classification. Figure \ref{fig:reweightedlayer} illustrates the ARL operation and Figure \ref{fig:attributeclass} shows how ARL is integrated into the attribute layers.

In summary, the motivation of designing the attribute re-weighting layer lies in two aspects. Firstly, by aggregating different kinds of attribute information to form a mid-level human semantic feature, we want to simulate the recognition process of human beings to form a compact attribute descriptor. Secondly, the detected object from object detection models can serve as a guidance for the global feature refinement process, promoting the performance of the recognition task.



\section{Experiments} \label{sec:experiments}
We evaluate our model on four widely used scene datasets including ADE20K \cite{Zhou2017ADE20KDataset}, MIT67 \cite{Quattoni2009MIT67Dataset}, SUN397 \cite{Xiao2010SunDatabase}, and Places365 \cite{Zhou2018PlacesDataset}.
We use models pre-trained on COCO Object and COCO Panoptic datasets to predict objects (i.e., \textit{things}) and concepts (i.e., \textit{stuff}) for every scene datasets. The resulting object predictions include all the $80$ objects from the COCO datasets and all the $91$ categories from the COCO Panoptic dataset with no overlaps.  
The final loss is then written as:
\begin{equation}\label{eq:finalloss}
  \mathcal{L}_{MASR} = \mathcal{L}_{att} + \mathcal{L}_{cls}
\end{equation}

\subsection{Datasets}

\textbf{ADE20K} \cite{Zhou2017ADE20KDataset} is a scene parsing dataset exhaustively annotated with objects. The dataset contains $20,210$ images for training, $2,000$ images for testing, $3,000$ images for validation and $1,055$ categories.

\textbf{MIT67} \cite{Quattoni2009MIT67Dataset} is a challenging indoor scene dataset containing $67$ categories and $15,620$ images. $80$ and $20$ images per category are used for training and testing, respectively.

\textbf{SUN397} \cite{Xiao2010SunDatabase} is a large-scale dataset containing $397$ scene categories and $108, 754$ images. It contains $50$ training images and $50$ testing images per category. 

\textbf{Places-365} dataset is the subset of the original Places database \cite{Zhou2018PlacesDataset} which includes over 7 million labeled pictures. 
The subset Places-365 contains $1,803,406$ training images and $365$ categories. 
The validation set contains $50$ images per category and the test set contains $900$ images per category.


\subsection{Implementation details}
In this section, we present the implementation details of our experiments. We use ResNet CNN models as backbones \cite{He2016ResNet} pre-trained on ImageNet \cite{Deng2009ImageNetDataset}. We add two fully connected layers $L^{\vert K \vert}$ and $L^{\vert A \vert}$ that projects $2048$-dimensional vectors into the category and attribute spaces, respectively. All the images are resized into a resolution of $256 \times 256$.
We first train our network on the scene dataset alone ($\mathcal{X}_I$) for an initial $15$ epochs in order to establish a basic feature representation. We then switch to the MASR setup and continue training for $85$ epochs using Equation \ref{eq:finalloss} loss. Our initial learning rate is set to $0.01$ for the classifier parameters and to $0.001$ for the base parameters and reduces by a factor of $0.1$ every $20$ epochs. We use a batchsize of $128$ and stochastic gradient descent optimizer. We set the threshold $\xi$ (see Eq. \ref{eq:discard} and \ref{eq:attributehat}) to $80\%$ and the minimum frequency $\beta$ to 20.




\begin{table*}
  \centering
  \caption{Ablation results for different backbones architectures.  Classification accuracy on MIT67 with resnet101.
Top k-accuracy and recall are shown. MASR (w/o ARL) denotes MASR without the attribute re-weighting layer}
  \label{tab:backbones}
  \resizebox{\textwidth}{!}{%
  \begin{tabular}{lcccccccccc}
    \toprule
    \multirow{2}{*}{Models} & \multicolumn{3}{c}{MIT67} & \multicolumn{3}{c}{SUN397} & \multicolumn{3}{c}{ADE20K}\\
    \cmidrule{2-10}
    & Top@1 & Top@2 & Top@5 & Top@1 & Top@2 & Top@5 & Top@1 & Top@2 & Top@5\\
    \midrule
    Baseline 1-VGG & 71.6 & 83.1 & 91.7 & 46.5 & 60.0 & 73.4 & 44.6 & 53.7 & 62.6 \\
    Baseline 1-DenseNet161 & 74.0 & 84.9 & 92.5 & 58.4 & 72.7 & 85.7 & 55.2 & 66.2 & 77.4\\
    Baseline 1-ResNet50  & 78.6 & 88.3 & 94.9 & 69.6 & 73.5 & 86.5 & 51.2 & 61.8 & 71.9 \\
    Baseline 1-ResNext101 & 82.3 & 90.3 & 96.0  & 70.8 &83.1 & 91.4 &54.7 & 65.6 & 74.8 \\

    \midrule
      MASR-VGG (w/o ARL) &  75.2 & 86.9 & 93.7 & 56.8 & 71.0 & 84.8 & 54.3 & 66.6 & 76.9 \\

      MASR-ResNet50 (w/o ARL) & 85.4 & 94.3 & 97.7  & 70.0 & 73.9 & 86.5 & 56.7 & 67.4 & 78.2 \\
      MASR-ResNext101 (w/o ARL)&  86.4 & 93.3 & 98.2 & 72.8 & 84.4 & 93.1 & 60.6 & 70.0 & 79.7\\
     \midrule
    MASR-VGG & 76.9 & 88.0 & 94.6 & 60.1 & 75.2 & 86.7 & 61.3 & 72.5 & 80.4  \\
    MASR-ResNet50 & 86.2 & 94.8 & 98.9 & 73.2 & 78.1 & 89.3 & 62.7 & 74.1 & 82.9  \\
    MASR-ResNext101 & 88.5 & 95.3 & 98.7 & 75.01 & 86.8& 94.6 & 64.4 & 75.2 & 85.1 \\
    \bottomrule
  \end{tabular}
  }
\end{table*}

\begin{table}[!htb]
\centering
\caption{Recognition accuracy comparison of some representative works on MIT67 dataset.} \label{tab:mit67}
  \begin{tabular}{lcc}
    \toprule
    Traditional methods & Venue  & Accuracy (\%) \\
     \midrule
     ROI \cite{Quattoni2009MIT67Dataset} & CVPR'09 & 26.05 \\
    CENTRIST \cite{Wu2011CENTRIST} & TPAMI'11 & 36.90 \\
    Hybrid parts \cite{Zheng2012Hybrid} & ECCV'12 & 39.80 \\
    BOP \cite{Juneja2013Blocks} & CVPR'13 & 46.10 \\
    GI ST + SP \cite{Zheng2012Hybrid} & ECCV'12 & 47.20 \\
    ISPR \cite{Lin2014SpatialPooling} & CVPR'14 & 50.10 \\
    Co-segmentation \cite{Sun2013Cosegmentation} & ICCV'13 & 51.40 \\
    DSFL \cite{Zuo2014Shareable} & ECCV'14 & 52.24 \\
    IFV \cite{Juneja2013Blocks} & CVPR'13 & 60.77 \\
    IFV + BOP \cite{Juneja2013Blocks} & CVPR'13 & 63.10 \\
    ISPR + IFV \cite{Lin2014SpatialPooling} & CVPR'14 & 68.50 \\
    \bottomrule
    \midrule
    CNN-based methods & Venue & Accuracy (\%) \\
    \midrule
    MOP-CNN \cite{Gong2014Orderless} & ECCV'14 & 68.90 \\
    HybridNet \cite{Zhou2014PlacesDataset} & NIPS'14 & 70.80 \\
    DSFL + CNN \cite{Zuo2014Shareable} & ECCV'14 & 76.23 \\
    DAG-CNN \cite{Yang2015DAGCNNs} & ICCV'15 &  77.50 \\
    Mix-CNN \cite{Shuqiang2019Codebook} & TOMM'19  & 79.63  \\
    CS(VGG-19) \cite{Xie2017HybridCNN} & TCSVT'17 & 82.24 \\
     LS-DHM \cite{Yang2015DAGCNNs} & ICCV'15 &    83.75  \\
      Multi-scale CNNs \cite{Herranz2016MultiScales} & CVPR'16 & 86.04 \\
    Dual CNN-DL \cite{Yang2018Dictionary} & AAAI'18 & 86.43  \\
     Multi-Resolution\cite{Wang2017MultiResolution} & TIP'17 & 86.7 \\
       SDO \cite{Cheng2018Objectness} & PR'18 & 86.76  \\
    SemanticAware \cite{Alejandro2020SemanticAware} & PR'20 & 87.10 \\
    \midrule
    MASR (Ours) & &  88.50 \\
    \bottomrule
  \end{tabular}
\end{table}

\begin{table}[!htb]
\centering
\caption{Comparison of our proposed approach with other methods on SUN397 dataset.} \label{tab:sun397}
  \begin{tabular}{llc}
    \toprule
     Traditional Methods & Venue & Accuracy (\%) \\
        \midrule
        Semantic Manifold \cite{Kwitt2012SemanticManifold} & ECCV'12 & 28.90 \\
        OTC \cite{Margolin2014OTC} & ECCV'14 & 34.56 \\
        cBoW + semantic \cite{Su2012SemanticAttributes} & IJCV'12 &35.60 \\
        Xiao et al \cite{Xiao2010SunDatabase} & CVPR'10 & 38.00 \\
        FV (SIFT + LCS) \cite{Sanchez2013FisherVector} & IJCV'13 &  47.20 \\
        OTC + HOG2 $\times$ 2 \cite{Margolin2014OTC} & ECCV'14 & 49.60 \\
    \toprule
        CNN-based methods & Venue & Accuracy (\%) \\
        \midrule
        DeCAF \cite {Donahue2014DeCAF} & ICML'14  & 40.94 \\
        MOP-CNN \cite{Gong2014Orderless} & ECCV'14 & 51.98 \\
        HybridNet \cite{Zhou2014PlacesDataset} & NIPS'14 & 53.86 \\
        Places-CNN \cite{Zhou2014PlacesDataset} & NIPS'14 & 54.23 \\
        Places-CNN ft \cite{Zhou2014PlacesDataset} & NIPS'14 & 56.20 \\
        DAG-CNN \cite{Yang2015DAGCNNs} & ICCV'15  & 56.20 \\
        Mix-CNN \cite{Shuqiang2019Codebook} & TOMM'19  & 57.47 \\
        CS(VGG-19) \cite{Xie2017HybridCNN} & TCSVT'17 & 64.53 \\
        LS-DHM \cite{Yang2015DAGCNNs} & ICCV'15 &     67.56 \\
        Dual CNN-DL \cite{Yang2018Dictionary} & AAAI'18 & 70.13 \\
        Multi-scale CNNs \cite{Herranz2016MultiScales} & CVPR'16  & 70.17\\ Multi-Resolution\cite{Wang2017MultiResolution} & TIP'17 & 72.0 \\

    SDO \cite{Cheng2018Objectness} & PR'18  &  73.41 \\
    SemanticAware \cite{Alejandro2020SemanticAware} & PR'20 & 74.04 \\
    \midrule
    MASR (Ours) &  & 75.01\\
    \bottomrule
  \end{tabular}
\end{table}

\begin{table}
  \centering
   \caption{Comparison results with other scene recognition methods on Places-365.} \label{tab:places365}
  \begin{tabular}{lcc}
    \toprule
    CNN-based methods & Venue & Accuracy (\%) \\
    \midrule
    Places365-ResNet \cite{Zhou2018PlacesDataset} & TPAMI'18 & 54.74 \\
    Places356-VGG \cite{Zhou2018PlacesDataset} & TPAMI'18 & 55.24 \\
    DenseNet-161  \cite{Alejandro2020SemanticAware} & PR'20 & 56.12\\
    Semantic Aware \cite{Alejandro2020SemanticAware} & PR'20 & 56.51 \\
    CNN-SMN \cite{Song2017MultiScale} & TIP'17 & 57.1 \\
    Fusing \cite{Sun2019Fusing} & TCSVT'18 & 57.27 \\
    Multi-Resolution\cite{Wang2017MultiResolution} & TIP'17 & 58.3 \\
    \midrule
    MASR (Ours) & & 56.61 \\
    \bottomrule
  \end{tabular}
  \end{table}
  
 \begin{table}
  \centering
  \caption{Comparison of scene recognition results on ADE20K dataset.} \label{tab:ade20k}
  \begin{tabular}{lcc}
    \toprule
    Methods & Venue & Accuracy (\%) \\
    \midrule
    ADE20k \cite{Zhou2017ADE20KDataset} & CVPR'17 & 45.38  \\
    Semantic Aware \cite{Alejandro2020SemanticAware} & PR'20 & 62.55 \\
    \midrule
    MASR (Ours)  & & 64.42\\
    \bottomrule
  \end{tabular}
\end{table}

\subsection{Comparison with the baselines}
Results on two datasets are shown in Table \ref{tab:backbones}. We observed that the classification baseline yields a reasonable performance by achieving a top@1 accuracy of $79.3\%$, $70.8\%$ and $54.7\%$ on MIT67, SUN397 and ADE20K datasets respectively.
We note that, for all the datasets, changing the backbone configurations from VGG to ResNet boost the performance accuracy by a factor of $7.7\%$ on MIT67 dataset.
Compared with the baselines, the use of attribute information alone (MASR w/o ARL) can improve the resulting top@1 performance by $4.1\%$ on MIT67, $2\%$ on SUN397, and $5.9\%$ on ADE20K datasets. Compared with the proposed framework, we boost the baseline performance by factor of $6\% - 10\%$ on all the datasets.

\subsection{Ablation studies}
We further investigate the discriminative ability of our framework by visualizing the learned embedding and analyzing the effectiveness of the ARL as follows:
\begin{figure}[h]
  \centering
  \includegraphics[width=\linewidth]{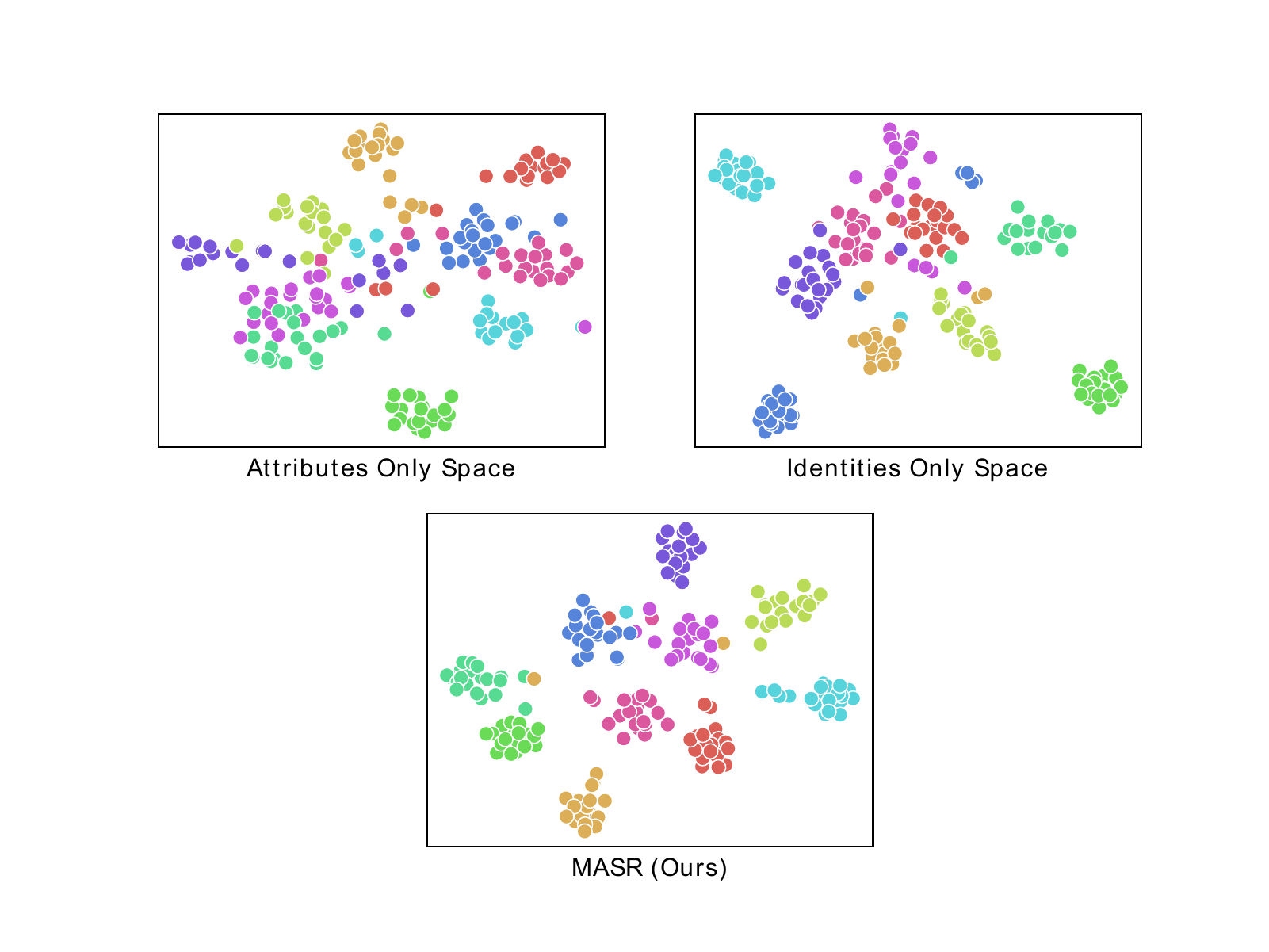}
  \caption{Feature distributions of 10 random categories in the three feature spaces. Images are from MIT67 and different colors represent different categories.}\label{fig:featurespace}
\end{figure}

\subsubsection{Feature visualization}
To understand the learned feature representation, we randomly selected $10$ categories from the test set and extracted their feature maps. We used t-SNE 
to visualize the embeddings by plotting their 2-dimension feature representation in Figure \ref{fig:featurespace}. Each point represents one image and points with the same color indicate image with the same category. In the attribute space, we see a somehow gathering of images with the same category while in the identity space, we observe a clear and constantly gathering of the images with the same color, which indicates that the model has learned a more discriminative feature representation. 
Compared to MARS feature space, 
the random selected categories are much more separated. We can see how images with the same color stay closer and far away from other color points. Only fewer different points are mixed. As a result, combing attributes with a re-weighting strategy helps to learn good representation and can achieve a better performance.

\subsubsection{The effectiveness of the attribute re-weighting layer}
To investigate the discriminative ability of the re-weighting layer, we preform further experiments using MASR without ARL (MARS w/o ARL) and report the results on Tables \ref{tab:backbones}. MASR (w/o ARL) is the network architecture where object attribute recognition task is simultaneously optimized with scene recognition task using Equations \ref{eq:attributehat} and \ref{eq:finalloss}. We activate the attribute loss contribution to the overall task only after the $15^{th}$ epoch. In general, we observe an increase in performance when adding attribute information to the Baseline 1 model. For example,
in MIT67 dataset, MASR (w/o ARL) achieves a top@1 accuracy of $86.4\%$ and outperforms Baseline 1 by a factor of $4.1\%$. Similarly, MASR, outperforms both MASR (w/o ARL) and Baseline 1 by a factor of $2.1\%$ and $6.2\%$ respectively. The performance improvement using ARL is consistent for all the datasets as shown in Table \ref{tab:backbones}. 
This demonstrates that the features of jointly optimizing attribute and category losses are consistently much better when attribute are re-weighted by their scores. Finding the right weight also 
enable MASR to achieve a more discriminative feature representation.

\subsection{Comparison with the state-of-the-art methods}

We compare MASR to a number of recent approaches and report the results on Tables \ref{tab:mit67},  \ref{tab:sun397}, \ref{tab:places365}, \ref{tab:ade20k}. In general, we outperform most of the recent works, including Semantic Aware \cite{Alejandro2020SemanticAware}, and achieve competitive results with  CNN-SMN \cite{Song2017MultiScale}, Fusing \cite{Sun2019Fusing} and Multi-Resolution\cite{Wang2017MultiResolution}. 

\textbf{Evaluation on MIT67 dataset}. Table \ref{tab:mit67} shows the evaluation results. We achieved an accuracy of $88.50\%$ and improved the baseline top@1 accuracy by a factor of $+6.2\%$ (from 82.3\% to 88.5\%). Our method largely outperforms traditional methods. For instance, we surpass ISPR + IFV \cite{Lin2014SpatialPooling} by $20\%$. In general, traditional methods based on descriptor such as GIST \cite{Laurent2005GIST} tends to perform poorly as they lack local structural information of a scene which is detrimental to the task.

\textbf{Evaluation on SUN397 dataset}. As shown in Table \ref{tab:sun397}, we achieved an accuracy of $75.01\%$. Similar to the evaluation on MIT67 dataset, we surpass the traditional methods by a large margin. We exceed OTC + HOG2 $\times$ 2 \cite{Margolin2014OTC} by $25.41\%$ and we gain a top@1 accuracy of $4.21\%$ over the baseline. Moreover, our results on this dataset surpasses Semantic Aware \cite{Alejandro2020SemanticAware} and Multi-Resolution \cite{Wang2017MultiResolution} by $0.97\%$ and $3.01\%$.

\textbf{Evaluation on Places-365 dataset}. The results on Table \ref{tab:places365} shows that Places-365 is a challenging dataset.  At it's the most diverse and largest dataset. 
Recent works such as CNN-SMN \cite{Song2017MultiScale} and Fusing \cite{Sun2019Fusing} which achieved a top@1 accuracy of $57.1\%$ and $57.27\%$, surpassed our model by small factors of $0.49\%$ and $0.66\%$, respectively. In addition, Multi-Resolution \cite{Wang2017MultiResolution}, which exploits knowledge computed on validation data outperforms our model by $1.69\%$. We note that Places-365 contains more than a million images and spans diverse scenes where common objects occur several times within the scene category with similar probability. 
Some discriminative parts of objets also appear with different shape and illumination. 
Consequently, the attribute objects obtained from this dataset by our technique is uniformly distributed across the scene categories.
 Our proposed technique may fail to take full advantage of the detected object scores. On small datasets, our technique performs well and can capture the discriminative object with high probabilities as shown on Table \ref{tab:backbones}. Nonetheless, we have achieved competitive results with other models and surpassed SemanticAware\cite{Alejandro2020SemanticAware}.

\textbf{Evaluation on ADE20K dataset}. On this dataset, not many results are reported as illustrated in Table \ref{tab:ade20k}. Yet, our method attains an accuracy of $64.42\%$ exceeding SemanticAware \cite{Alejandro2020SemanticAware} by $1.87\%$. We also improved the baseline performance from $54.7\%$ to $64.4\%$ and particularly outperformed the initial scene parsing model \cite{Zhou2017ADE20KDataset} on ADE20K datasets by $19.04\%$.

\subsection{Evaluation of attribute recognition task}
\begin{table*}
  \centering
    \caption{Attribute recognition accuracy on MIT67. We only show 10 attribute objects. AP stands for Average Precision. B2 denotes 'Baseline 2'.}
 \label{tab:attributemit67}
    \resizebox{\textwidth}{!}{%
  \begin{tabular}{lcccccccccccc}
    \toprule
     &person  & counter & floor & shelf &window&
      ceiling & cabinet &table &building & rug & AP \\
    \midrule

    B2 & 75.5  &  89.3 & 77.7 & 77.4 & 76.2 &91.5&72.0& 76.7 &
         84.0 & 81.4 & 86.0 \\
    MASR & 72.7& 100.0& 73.0 &77.0 & 90.2& 88.9 & 89.6 &100.0 & 87.5 & 80.0& 87.1  \\
    \bottomrule
  \end{tabular}
  }
\end{table*}
\begin{table*}
  \caption{Attribute recognition accuracy on SUN397. We also report the 10 most occurring attributes. PAV: pavement, AP: average precision, MOU: mountain, BUI: building, CA: cabinet, WI: Windows, CE: ceiling. B2 denotes 'Baseline 2'}
 \label{tab:attributesun397}
  \centering
      \resizebox{\textwidth}{!}{%
  \begin{tabular}{lccccccccccccccc}
    \toprule
    & person &river & sea & shelf &  WI  & tree  & CE &
    CA & PAV & MOU & grass & dirt & BUI & AP \\
    \midrule
    B2 &71.6& 97.7 & 98.2 & 89.5 &75.8 & 93.2 &  90.9  &  97.0 &
    74.9 &  92.2 & 90.3 &  87.8 & 84.7  & 88.1\\

    MASR & 77.4&93.9 &93.0 & 82.6 & 79.7 &
      92.5 & 87.1 & 92.5 & 76.2 & 88.0 & 89.3 & 88.3 & 83.0 & 89.7 \\
    \bottomrule
  \end{tabular}
  }
\end{table*}

 In this work, we developed a scene recognition approach which leverages information contained in automatically detected attributes to improve its classification results. However, we believe that these detected attributes can also be used for attribute recognition task. We test attribute recognition on MIT67 and SUN397 datasets and report the attribute detection score in Tables \ref{tab:attributemit67}, \ref{tab:attributesun397}. We use the Baseline 2, trained specifically for the attribute recognition task (See Eq. \ref{eq:baseline2loss}). The Baseline 2 predicts a set of attribute given a scene image. We compute the precision as the ratio $\frac{TP}{TP + FP}$ where $TP$ is the number of true positives and $FP$ the number of false positives. 
 By comparing the results of MASR and Baseline 2, we can drawn two conclusions: 
 First, on all datasets, the overall attribute recognition accuracy is improved by the proposed MASR network to some extent. The improvements are $1.1\%$ and $0.08\%$ on MIT67 and SUN397 respectively. In a nutshell, the integration of classification introduces some degree of complementary information and helps in learning a more discriminative attribute recognition model. 
 Secondly, we observe that the recognition rate of some attributes decreases for MASR, such as ``person'', ``floor'', and ``ceiling'' in MIT67 dataset. This can be explained by the fact that MIT67 is an indoor dataset. As a results, attributes such as ``floor'', ``ceiling'', span several scene categories, and are present with a high probability across various different categories. Moreover, 
 the reason probably lies in the multi-task nature of MASR. The model is primarily optimized for classification while attribute recognition task is a multi-label classification problem. Moreover, ambiguous images of certain attributes may be incorrectly predicted by the object detection model.  Nevertheless, the improvement on the two datasets is still encouraging and further investigations should be critical.


\section{Conclusion and future work} \label{sec:conclusion}

In this paper, we proposed a Multi-task Attribute-scene recognition (MASR) network that exploits both category labels and attribute annotations. By combining the two tasks, the MASR network is able to learn more discriminative feature representations for scene, including attribute features and scene features. Specifically, we mined attribute labels from existing object-centric datasets and considered their predictions as additional cues for scene classification.
Extensive experiment on four large-scale datasets showed that our method achieves competitive recognition accuracy compared to the state-of-the-art methods. We also showed that the proposed MASR yields improvement in the attribute recognition task over the baseline in all the testing datasets.

In future work, we will investigate the transferability and scalability of scene attributes. For example, we could adapt the attribute model learned on SUN397 dataset to other scene datasets. Secondly, attributes provide a bridge to the image-text understanding. We could investigate a system using attributes to retrieve the relevant scene images. It is useful in solving specific image retrieval problems, in which the query image is missing and can be described by attributes.

\bibliographystyle{IEEEtran}
\bibliography{egbib}

\begin{thebibliography}{10}
\providecommand{\url}[1]{#1}
\csname url@samestyle\endcsname
\providecommand{\newblock}{\relax}
\providecommand{\bibinfo}[2]{#2}
\providecommand{\BIBentrySTDinterwordspacing}{\spaceskip=0pt\relax}
\providecommand{\BIBentryALTinterwordstretchfactor}{4}
\providecommand{\BIBentryALTinterwordspacing}{\spaceskip=\fontdimen2\font plus
\BIBentryALTinterwordstretchfactor\fontdimen3\font minus
  \fontdimen4\font\relax}
\providecommand{\BIBforeignlanguage}[2]{{%
\expandafter\ifx\csname l@#1\endcsname\relax
\typeout{** WARNING: IEEEtran.bst: No hyphenation pattern has been}%
\typeout{** loaded for the language `#1'. Using the pattern for}%
\typeout{** the default language instead.}%
\else
\language=\csname l@#1\endcsname
\fi
#2}}
\providecommand{\BIBdecl}{\relax}
\BIBdecl

\bibitem{Cordts2016CityscapeDataset}
M.~{Cordts}, M.~{Omran}, S.~{Ramos}, T.~{Rehfeld}, M.~{Enzweiler},
  R.~{Benenson}, U.~{Franke}, S.~{Roth}, and B.~{Schiele}, ``The cityscapes
  dataset for semantic urban scene understanding,'' in \emph{2016 IEEE CVPR},
  2016, pp. 3213--3223.

\bibitem{Zhou2014PlacesDataset}
B.~Zhou, A.~Lapedriza, J.~Xiao, A.~Torralba, and A.~Oliva, ``Learning deep
  features for scene recognition using places database,'' in \emph{Proceedings
  of the 27th NIPS - Volume 1}, ser. NIPS'14.\hskip 1em plus 0.5em minus
  0.4em\relax Cambridge, MA, USA: MIT Press, 2014, p. 487–495.

\bibitem{Alejandro2020SemanticAware}
A.~López-Cifuentes, M.~Escudero-Viñolo, J.~Bescós, and Álvaro
  García-Martín, ``Semantic-aware scene recognition,'' \emph{Pattern
  Recognition}, vol. 102, p. 107256, 2020.

\bibitem{Wu2019detectron2}
Y.~Wu, A.~Kirillov, F.~Massa, W.-Y. Lo, and R.~Girshick, ``Detectron2,''
  \url{https://github.com/facebookresearch/detectron2}, 2019.

\bibitem{Zhou2017ADE20KDataset}
B.~{Zhou}, H.~{Zhao}, X.~{Puig}, S.~{Fidler}, A.~{Barriuso}, and A.~{Torralba},
  ``Scene parsing through ade20k dataset,'' in \emph{2017 IEEE Conference on
  Computer Vision and Pattern Recognition (CVPR)}, 2017, pp. 5122--5130.

\bibitem{Xiao2010SunDatabase}
J.~{Xiao}, J.~{Hays}, K.~A. {Ehinger}, A.~{Oliva}, and A.~{Torralba}, ``Sun
  database: Large-scale scene recognition from abbey to zoo,'' in \emph{2010
  IEEE Computer Society Conference on Computer Vision and Pattern Recognition},
  2010, pp. 3485--3492.

\bibitem{Dalal2005HOG}
N.~{Dalal} and B.~{Triggs}, ``Histograms of oriented gradients for human
  detection,'' in \emph{2005 IEEE Computer Society Conference on Computer
  Vision and Pattern Recognition (CVPR'05)}, vol.~1, 2005, pp. 886--893 vol. 1.

\bibitem{Lowe1999SIFT}
D.~G. {Lowe}, ``Object recognition from local scale-invariant features,'' in
  \emph{Proceedings of the Seventh IEEE International Conference on Computer
  Vision}, vol.~2, 1999, pp. 1150--1157 vol.2.

\bibitem{Shechtman2007SSIM}
E.~{Shechtman} and M.~{Irani}, ``Matching local self-similarities across images
  and videos,'' in \emph{2007 IEEE Conference on Computer Vision and Pattern
  Recognition}, 2007, pp. 1--8.

\bibitem{Meng2012SceneRecognition}
X.~Meng, Z.~Wang, and L.~Wu, ``Building global image features for scene
  recognition,'' \emph{Pattern Recognition}, vol.~45, no.~1, pp. 373 -- 380,
  2012.

\bibitem{Cimpoi2015DeepFilter}
M.~{Cimpoi}, S.~{Maji}, and A.~{Vedaldi}, ``Deep filter banks for texture
  recognition and segmentation,'' in \emph{2015 IEEE Conference on Computer
  Vision and Pattern Recognition (CVPR)}, 2015, pp. 3828--3836.

\bibitem{Donahue2014DeCAF}
J.~Donahue, Y.~Jia, O.~Vinyals, J.~Hoffman, N.~Zhang, E.~Tzeng, and T.~Darrell,
  ``Decaf: A deep convolutional activation feature for generic visual
  recognition,'' in \emph{Proceedings of the 31st International Conference on
  Machine Learning - Volume 32}, ser. ICML'14.\hskip 1em plus 0.5em minus
  0.4em\relax JMLR.org, 2014, p. I–647–I–655.

\bibitem{Xie2017HybridCNN}
G.~{Xie}, X.~{Zhang}, S.~{Yan}, and C.~{Liu}, ``Hybrid cnn and dictionary-based
  models for scene recognition and domain adaptation,'' \emph{IEEE Transactions
  on Circuits and Systems for Video Technology}, vol.~27, no.~6, pp.
  1263--1274, 2017.

\bibitem{Chen2020PrototypeAgnostic}
G.~{Chen}, X.~{Song}, H.~{Zeng}, and S.~{Jiang}, ``Scene recognition with
  prototype-agnostic scene layout,'' \emph{IEEE Transactions on Image
  Processing}, vol.~29, pp. 5877--5888, 2020.

\bibitem{Zamir2018Taskonomy}
A.~R. {Zamir}, A.~{Sax}, W.~{Shen}, L.~{Guibas}, J.~{Malik}, and S.~{Savarese},
  ``Taskonomy: Disentangling task transfer learning,'' in \emph{2018 IEEE/CVF
  Conference on Computer Vision and Pattern Recognition}, 2018, pp. 3712--3722.

\bibitem{Quattoni2009MIT67Dataset}
A.~{Quattoni} and A.~{Torralba}, ``Recognizing indoor scenes,'' in \emph{2009
  IEEE Conference on Computer Vision and Pattern Recognition}, 2009, pp.
  413--420.

\bibitem{Zhou2018PlacesDataset}
B.~{Zhou}, A.~{Lapedriza}, A.~{Khosla}, A.~{Oliva}, and A.~{Torralba},
  ``Places: A 10 million image database for scene recognition,'' \emph{IEEE
  Transactions on Pattern Analysis and Machine Intelligence}, vol.~40, no.~6,
  pp. 1452--1464, 2018.

\bibitem{Laurent2005GIST}
A.~Oliva, ``Chapter 41 - gist of the scene,'' in \emph{Neurobiology of
  Attention}, L.~Itti, G.~Rees, and J.~K. Tsotsos, Eds.\hskip 1em plus 0.5em
  minus 0.4em\relax Burlington: Academic Press, 2005, pp. 251 -- 256.

\bibitem{Lazebnik2006BeyondBags}
S.~{Lazebnik}, C.~{Schmid}, and J.~{Ponce}, ``Beyond bags of features: Spatial
  pyramid matching for recognizing natural scene categories,'' in \emph{2006
  IEEE Computer Society Conference on Computer Vision and Pattern Recognition
  (CVPR'06)}, vol.~2, 2006, pp. 2169--2178.

\bibitem{Krapac2011Modeling}
J.~{Krapac}, J.~{Verbeek}, and F.~{Jurie}, ``Modeling spatial layout with
  fisher vectors for image categorization,'' in \emph{2011 International
  Conference on Computer Vision}, 2011, pp. 1487--1494.

\bibitem{Wu2011CENTRIST}
J.~{Wu} and J.~M. {Rehg}, ``Centrist: A visual descriptor for scene
  categorization,'' \emph{IEEE Transactions on Pattern Analysis and Machine
  Intelligence}, vol.~33, no.~8, pp. 1489--1501, 2011.

\bibitem{Margolin2014OTC}
R.~Margolin, L.~Zelnik-Manor, and A.~Tal, ``Otc: A novel local descriptor for
  scene classification,'' in \emph{Computer Vision -- ECCV 2014}.\hskip 1em
  plus 0.5em minus 0.4em\relax Cham: Springer International Publishing, 2014,
  pp. 377--391.

\bibitem{Shuqiang2019Codebook}
S.~Jiang, G.~Chen, X.~Song, and L.~Liu, ``Deep patch representations with
  shared codebook for scene classification,'' \emph{ACM Trans. Multimedia
  Comput. Commun. Appl.}, vol.~15, no.~1s, Jan. 2019.

\bibitem{Gong2014Orderless}
Y.~Gong, L.~Wang, R.~Guo, and S.~Lazebnik, ``Multi-scale orderless pooling of
  deep convolutional activation features,'' in \emph{Computer Vision -- ECCV
  2014}.\hskip 1em plus 0.5em minus 0.4em\relax Cham: Springer International
  Publishing, 2014, pp. 392--407.

\bibitem{Yang2015DAGCNNs}
S.~{Yang} and D.~{Ramanan}, ``Multi-scale recognition with dag-cnns,'' in
  \emph{2015 IEEE International Conference on Computer Vision (ICCV)}, 2015,
  pp. 1215--1223.

\bibitem{Cheng2018Objectness}
X.~Cheng, J.~Lu, J.~Feng, B.~Yuan, and J.~Zhou, ``Scene recognition with
  objectness,'' \emph{Pattern Recognition}, vol.~74, pp. 474 -- 487, 2018.

\bibitem{Herranz2016MultiScales}
L.~{Herranz}, S.~{Jiang}, and X.~{Li}, ``Scene recognition with cnns: Objects,
  scales and dataset bias,'' in \emph{2016 IEEE Conference on Computer Vision
  and Pattern Recognition (CVPR)}, 2016, pp. 571--579.

\bibitem{Yang2018Dictionary}
Y.~Liu, Q.~Chen, W.~Chen, and I.~Wassell, ``Dictionary learning inspired deep
  network for scene recognition,'' 2018.

\bibitem{Pei2021PlacePerception}
P.~Li, X.~Li, X.~Li, H.~Pan, M.~Khyam, M.~Noor-A-Rahim, and S.~S. Ge, ``Place
  perception from the fusion of different image representation,'' \emph{Pattern
  Recognition}, vol. 110, p. 107680, 2021.

\bibitem{Lin2019AttributeReID}
Y.~Lin, L.~Zheng, Z.~Zheng, Y.~Wu, Z.~Hu, C.~Yan, and Y.~Yang, ``Improving
  person re-identification by attribute and identity learning,'' \emph{Pattern
  Recognition}, vol.~95, pp. 151 -- 161, 2019.

\bibitem{Tay2019AANetReID}
C.~{Tay}, S.~{Roy}, and K.~{Yap}, ``Aanet: Attribute attention network for
  person re-identifications,'' in \emph{2019 IEEE/CVF Conference on Computer
  Vision and Pattern Recognition (CVPR)}, 2019, pp. 7127--7136.

\bibitem{Yao2011ActionRecognition}
B.~{Yao}, X.~{Jiang}, A.~{Khosla}, A.~L. {Lin}, L.~{Guibas}, and L.~{Fei-Fei},
  ``Human action recognition by learning bases of action attributes and
  parts,'' in \emph{2011 International Conference on Computer Vision}, 2011,
  pp. 1331--1338.

\bibitem{Roy2019UniversalAttribute}
D.~{Roy}, K.~S.~R. {Murty}, and C.~K. {Mohan}, ``Unsupervised universal
  attribute modeling for action recognition,'' \emph{IEEE Transactions on
  Multimedia}, vol.~21, no.~7, pp. 1672--1680, 2019.

\bibitem{Ferrari2007VisualAttributes}
V.~Ferrari and A.~Zisserman, ``Learning visual attributes,'' in
  \emph{Proceedings of the 20th International Conference on Neural Information
  Processing Systems}, ser. NIPS'07.\hskip 1em plus 0.5em minus 0.4em\relax Red
  Hook, NY, USA: Curran Associates Inc., 2007, p. 433–440.

\bibitem{Lampert2009AttributeTransfer}
C.~H. {Lampert}, H.~{Nickisch}, and S.~{Harmeling}, ``Learning to detect unseen
  object classes by between-class attribute transfer,'' in \emph{2009 IEEE
  Conference on Computer Vision and Pattern Recognition}, 2009, pp. 951--958.

\bibitem{Zeng2020SceneAttribute}
H.~{Zeng}, X.~{Song}, G.~{Chen}, and S.~{Jiang}, ``Learning scene attribute for
  scene recognition,'' \emph{IEEE Transactions on Multimedia}, vol.~22, no.~6,
  pp. 1519--1530, 2020.

\bibitem{Patterson2012SUN}
G.~{Patterson} and J.~{Hays}, ``Sun attribute database: Discovering,
  annotating, and recognizing scene attributes,'' in \emph{2012 IEEE Conference
  on Computer Vision and Pattern Recognition}, 2012, pp. 2751--2758.

\bibitem{Rueda2018HumanActivity}
F.~M. {Rueda} and G.~A. {Fink}, ``Learning attribute representation for human
  activity recognition,'' in \emph{2018 24th International Conference on
  Pattern Recognition (ICPR)}, 2018, pp. 523--528.

\bibitem{Xian2018AWA2Dataset}
Y.~{Xian}, C.~H. {Lampert}, B.~{Schiele}, and Z.~{Akata}, ``Zero-shot
  learning—a comprehensive evaluation of the good, the bad and the ugly,''
  \emph{IEEE Transactions on Pattern Analysis and Machine Intelligence},
  vol.~41, no.~9, pp. 2251--2265, 2019.

\bibitem{Olivia2001Holistic}
A.~Oliva and A.~Torralba, ``Modeling the shape of the scene: A holistic
  representation of the spatial envelope,'' \emph{International Journal of
  Computer Vision}, vol.~42, no.~3, pp. 145--175, May 2001.

\bibitem{Wang2013AttributeLocalization}
S.~{Wang}, J.~{Joo}, Y.~{Wang}, and S.~{Zhu}, ``Weakly supervised learning for
  attribute localization in outdoor scenes,'' in \emph{2013 IEEE Conference on
  Computer Vision and Pattern Recognition}, 2013, pp. 3111--3118.

\bibitem{Ouyang2015AttributeImageNet}
W.~{Ouyang}, H.~{Li}, X.~{Zeng}, and X.~{Wang}, ``Learning deep representation
  with large-scale attributes,'' in \emph{2015 IEEE International Conference on
  Computer Vision (ICCV)}, 2015, pp. 1895--1903.

\bibitem{Patterson2014BeyondCategories}
G.~Patterson, C.~Xu, H.~Su, and J.~Hays, ``The sun attribute database: Beyond
  categories for deeper scene understanding,'' \emph{International Journal of
  Computer Vision}, vol. 108, no.~1, p.~59, May 2014.

\bibitem{He2016ResNet}
K.~{He}, X.~{Zhang}, S.~{Ren}, and J.~{Sun}, ``Deep residual learning for image
  recognition,'' in \emph{2016 IEEE Conference on Computer Vision and Pattern
  Recognition (CVPR)}, 2016, pp. 770--778.

\bibitem{Deng2009ImageNetDataset}
J.~{Deng}, W.~{Dong}, R.~{Socher}, L.~{Li}, {Kai Li}, and {Li Fei-Fei},
  ``Imagenet: A large-scale hierarchical image database,'' in \emph{2009 IEEE
  Conference on Computer Vision and Pattern Recognition}, 2009, pp. 248--255.

\bibitem{Zheng2012Hybrid}
Y.~Zheng, Y.-G. Jiang, and X.~Xue, ``Learning hybrid part filters for scene
  recognition,'' in \emph{Computer Vision -- ECCV 2012}.\hskip 1em plus 0.5em
  minus 0.4em\relax Berlin, Heidelberg: Springer Berlin Heidelberg, 2012, pp.
  172--185.

\bibitem{Juneja2013Blocks}
M.~{Juneja}, A.~{Vedaldi}, C.~V. {Jawahar}, and A.~{Zisserman}, ``Blocks that
  shout: Distinctive parts for scene classification,'' in \emph{2013 IEEE
  Conference on Computer Vision and Pattern Recognition}, 2013, pp. 923--930.

\bibitem{Lin2014SpatialPooling}
D.~{Lin}, C.~{Lu}, R.~{Liao}, and J.~{Jia}, ``Learning important spatial
  pooling regions for scene classification,'' in \emph{2014 IEEE Conference on
  Computer Vision and Pattern Recognition}, 2014, pp. 3726--3733.

\bibitem{Sun2013Cosegmentation}
J.~{Sun} and J.~{Ponce}, ``Learning discriminative part detectors for image
  classification and cosegmentation,'' in \emph{2013 IEEE International
  Conference on Computer Vision}, 2013, pp. 3400--3407.

\bibitem{Zuo2014Shareable}
Z.~Zuo, G.~Wang, B.~Shuai, L.~Zhao, Q.~Yang, and X.~Jiang, ``Learning
  discriminative and shareable features for scene classification,'' in
  \emph{Computer Vision -- ECCV 2014}.\hskip 1em plus 0.5em minus 0.4em\relax
  Cham: Springer International Publishing, 2014, pp. 552--568.

\bibitem{Wang2017MultiResolution}
L.~{Wang}, S.~{Guo}, W.~{Huang}, Y.~{Xiong}, and Y.~{Qiao}, ``Knowledge guided
  disambiguation for large-scale scene classification with multi-resolution
  cnns,'' \emph{IEEE Transactions on Image Processing}, vol.~26, no.~4, pp.
  2055--2068, 2017.

\bibitem{Kwitt2012SemanticManifold}
R.~Kwitt, N.~Vasconcelos, and N.~Rasiwasia, ``Scene recognition on the semantic
  manifold,'' in \emph{Computer Vision -- ECCV 2012}.\hskip 1em plus 0.5em
  minus 0.4em\relax Berlin, Heidelberg: Springer Berlin Heidelberg, 2012, pp.
  359--372.

\bibitem{Su2012SemanticAttributes}
Y.~Su and F.~Jurie, ``Improving image classification using semantic
  attributes,'' \emph{International Journal of Computer Vision}, vol. 100,
  no.~1, pp. 59--77, may 2012.

\bibitem{Sanchez2013FisherVector}
J.~S{\'{a}}nchez, F.~Perronnin, T.~Mensink, and J.~Verbeek, ``Image
  classification with the fisher vector: Theory and practice,''
  \emph{International Journal of Computer Vision}, vol. 105, no.~3, pp.
  222--245, jun 2013.

\bibitem{Song2017MultiScale}
X.~{Song}, S.~{Jiang}, and L.~{Herranz}, ``Multi-scale multi-feature context
  modeling for scene recognition in the semantic manifold,'' \emph{IEEE
  Transactions on Image Processing}, vol.~26, no.~6, pp. 2721--2735, 2017.

\bibitem{Sun2019Fusing}
N.~{Sun}, W.~{Li}, J.~{Liu}, G.~{Han}, and C.~{Wu}, ``Fusing object semantics
  and deep appearance features for scene recognition,'' \emph{IEEE Transactions
  on Circuits and Systems for Video Technology}, vol.~29, no.~6, pp.
  1715--1728, 2019.

\end{thebibliography}







\end{document}